\documentclass[10pt]{article} 

\usepackage[accepted]{rlj} 

\usepackage{amssymb}            
\usepackage{mathtools}          
\usepackage{mathrsfs}           
\usepackage{graphicx}           
\usepackage[space]{grffile}     
\usepackage{url}                
\usepackage{lipsum}             
\usepackage{subfig}
\usepackage{wrapfig}
\usepackage{cleveref}
\usepackage{dsfont}
\usepackage{algorithm}
\usepackage{algpseudocode}
\usepackage{booktabs}
\usepackage{array}
\usepackage[rightcaption]{sidecap}
\hypersetup{
    citecolor=linkblue,
    linkcolor=linkblue,
    urlcolor=linkblue
}

\title{Trajectory First:\\A Curriculum for Discovering Diverse Policies}
\setrunningtitle{Trajectory First: A Curriculum for Discovering Diverse Policies}

\author{Cornelius V.~Braun\textsuperscript{1}, Sayantan Auddy\textsuperscript{1}, Marc Toussaint\textsuperscript{1, 2}}

\emails{braun@tu-berlin.de}

\affiliations{
$^{1}$\textbf{TU Berlin}\\
$^{2}$\textbf{Robotics Institute Germany}\\
}

\contribution{
    This paper studies the problem of optimality-constrained training of diverse policies in the context of robot manipulation. 
    We empirically show that collision penalties and contact mode switching are key challenges for existing methods, leading to low diversity.
}
{
    Constrained diversity optimization has become a popular framework for training a set of diverse policies~\citep{zahavy2023discovering}.
    Previous work considers predominantly free-space locomotion and navigation tasks~\citep{zahavy2023discovering, vlastelica2024offline, pang2023global}.
    In robot manipulation, however, different solutions, such as differing grasp orientations or contact points, tend to lie in different modes on the reward manifold, so they cannot be bridged with local exploration~\citep{choi2024unsupervised}.
    While these observations have been discussed in the context of trajectory optimization before~\citep{pang2023global}, we demonstrate that they also limit diversity-focused actor-critic methods.
}
\contribution{
    We introduce an efficient two-stage \textit{trajectory-first} curriculum tailored to constrained diversity optimization in RL.
}
{
    The algorithm first explores smooth trajectories using a novel blackbox optimization method and uses the discovered behaviors to anchor policy training.
    This exploration strategy, while simple, is a surprisingly efficient recipe for training diverse robot manipulation skills.
}
\contribution{
    We present an algorithmic recipe for diversity-preserving offline-to-online RL.
    In particular, we discover specific components to efficiently train diverse policies from offline data.
}
{
    Prior offline-to-online RL methods~\citep{ball2023efficient, zhou2025efficient} focus on the training of single policies.
    While our curriculum generates the exploration data autonomously, our findings on offline-to-online RL in the context of diversity-seeking RL are applicable to other data sources beyond trajectory optimization data, e.g., expert demonstrations.
}


\newcommand{\myAbs}{%
Being able to solve a task in diverse ways makes agents more robust to task variations and less prone to local optima.
In this context, constrained diversity optimization has become a useful reinforcement learning (RL) framework for training a set of diverse agents in parallel.
However, existing constrained-diversity RL methods often under-explore in complex tasks such as robot manipulation, resulting in limited behavioral diversity. 
We address this with a two-stage curriculum that introduces a spline-based trajectory prior as an inductive bias to produce diverse, high-reward behaviors in an initial stage, and then distills these behaviors into reactive, step-wise policies in a second stage.
In our empirical evaluation, we provide novel insights into challenges of diversity-targeted training and show that our curriculum increases the diversity of learned skills while maintaining high task performance.%
}

\summary{\protect\myAbs}

  \renewcommand{\a}{\alpha}
  \renewcommand{\b}{\beta}
  \renewcommand{\d}{\delta}

  \renewcommand{\l}{\lambda}
  \renewcommand{\L}{\Lambda}
    \newcommand{\m}{\mu}

  \renewcommand{\k}{\kappa}
  \renewcommand{\o}{\omega}
  \renewcommand{\O}{\Omega}

  \renewcommand{\r}{\varrho}
    \newcommand{\s}{\sigma}
  
  \renewcommand{\t}{\theta}

  \renewcommand{\AA}{{\cal A}}

    \newcommand{\DD}{{\cal D}}

    \newcommand{\HH}{{\cal H}}

  \renewcommand{\SS}{{\cal S}}

  \newcommand{\RRR}{{\mathbb{R}}}

  \newcommand{\vx}{\mathbf{x}}

  \renewcommand{\[}{\Big[}

  \renewcommand{\|}{\,|\,}
  
  \renewcommand{\=}{\!=\!}

  \newcommand{\tsum}{\textstyle\sum}
  \newcommand{\st}{~~\text{s.t.}~~}

  \renewcommand{\vec}{\boldsymbol}

  \providecommand{\href}[2]{{\color{blue}USE PDFLATEX!}}
  \providecommand{\url}[2]{\href{#1}{{\color{blue}#2}}}

  \newcommand{\fullhide}[1]{\message{^^JHIDE--Warning (hidden)!^^J}}
  \newcommand{\mytitle}[1]{\title{#1}\newcommand{\thetitle}{#1}}
  \newcommand{\header}{\begin{document}\mytitle\cleardefs}
  \newcommand{\contents}{{\tableofcontents}\renewcommand{\contents}{}}
  
  \newcommand{\widepaper}{\usepackage{geometry}\geometry{a4paper,hdivide={25mm,*,25mm},vdivide={25mm,*,25mm}}}
  \newcommand{\moviex}[2]{\movie[externalviewer]{#1}{#2}} 
  \newcommand{\rbox}[1]{\fboxrule2mm\fcolorbox[rgb]{1,.85,.85}{1,.85,.85}{#1}}
  \newcommand{\mpage}[2]{{\begin{minipage}{#1\columnwidth}#2\end{minipage}}}
  \newcommand{\redbox}[2]{\fboxrule1mm\fcolorbox[rgb]{1,.7,.7}{1,.7,.7}{\begin{minipage}{#1\columnwidth}\center#2\end{minipage}}}
  \newcommand{\onecol}[2]{
    \begin{minipage}[c]{#1\columnwidth}#2\end{minipage}}
  \newcommand{\twocol}[5][0]{
    \begin{minipage}[c]{#2\columnwidth}#4\end{minipage}\hspace*{#1\columnwidth}%
    \begin{minipage}[c]{#3\columnwidth}#5\end{minipage}}
  \newcommand{\threecol}[7][0]{%
    \begin{minipage}[c]{#2\columnwidth}#5\end{minipage}\hspace*{#1\columnwidth}%
    \begin{minipage}[c]{#3\columnwidth}#6\end{minipage}\hspace*{#1\columnwidth}%
    \begin{minipage}[c]{#4\columnwidth}#7\end{minipage}}
  \newcommand{\threecoltext}[7][c]{
    \begin{minipage}[#1]{#2\textwidth}#5\end{minipage}%
    \begin{minipage}[#1]{#3\textwidth}#6\end{minipage}%
    \begin{minipage}[#1]{#4\textwidth}#7\end{minipage}}
  \newcommand{\threecoltop}[7][0]{%
   \begin{minipage}[t]{#2\columnwidth}#5\end{minipage}\hspace*{#1\columnwidth}%
   \begin{minipage}[t]{#3\columnwidth}#6\end{minipage}\hspace*{#1\columnwidth}%
   \begin{minipage}[t]{#4\columnwidth}#7\end{minipage}}
  \newcommand{\fourcol}[9][0]{%
   \begin{minipage}[c]{#2\columnwidth}#6\end{minipage}\hspace*{#1\columnwidth}%
   \begin{minipage}[c]{#3\columnwidth}#7\end{minipage}\hspace*{#1\columnwidth}%
   \begin{minipage}[c]{#4\columnwidth}#8\end{minipage}\hspace*{#1\columnwidth}%
   \begin{minipage}[c]{#5\columnwidth}#9\end{minipage}}
  \newcommand{\helvetica}[4]{\setlength{\unitlength}{1pt}\fontsize{#1}{#1}\linespread{#2}\usefont{OT1}{phv}{#3}{#4}}
  \newcommand{\helve}[1]{\helvetica{#1}{1.5}{m}{n}}
  \newcommand{\german}{\usepackage[german]{babel}\usepackage[utf8]{inputenc}}
\newcommand{\sans}{\renewcommand{\familydefault}{\sfdefault}}

\newcommand{\norm}[1]{|\!|#1|\!|}
\newcommand{\expr}[1]{[\hspace{-.2ex}[#1]\hspace{-.2ex}]}

\newcommand{\Jwi}{J^\sharp_W}
\newcommand{\THi}{T^\sharp_H}
\newcommand{\Jci}{J^\natural_C}
\newcommand{\hJi}{{\bar J}^\sharp}
\renewcommand{\|}{\,|\,}
\renewcommand{\=}{\!=\!}
\newcommand{\myminus}{{\hspace*{-.0pt}\text{\rm -}\hspace*{-.5pt}}}
\newcommand{\myplus}{{\hspace*{-.0pt}\text{\rm +}\hspace*{-.5pt}}}
\newcommand{\1}{{\myminus1}}
\newcommand{\2}{{\myminus2}}
\newcommand{\3}{{\myminus3}}
\newcommand{\mT}{{\text{\rm -}\hspace*{-1pt}\top}}
\newcommand{\po}{{\myplus1}}
\newcommand{\pt}{{\myplus2}}
\newcommand{\T}{{\!\top\!}}
\newcommand{\xT}{{\underline x}}
\newcommand{\uT}{{\underline u}}
\newcommand{\zT}{{\underline z}}
\newcommand{\Sum}{\textstyle\sum}
\newcommand{\Int}{\textstyle\int}
\newcommand{\Prod}{\textstyle\prod}

\newenvironment{centy}{
\vspace{15mm}
\large
\hspace*{5mm}
\begin{minipage}{8cm}\it\color{blue}
}{
\end{minipage}
}

\newcommand{\old}{{\text{old}}}
\newcommand{\MAP}{{\text{MAP}}}
\newcommand{\ML}{{\text{ML}}}

\newcommand{\pub}[1]{{\color{green}\helvetica{8}{1.3}{m}{n}#1\\}}
\newcommand{\KL}[2]{\opKL\big(#1\,\big|\!\big|\,#2\big)} 

\renewcommand{\show}[2][.7]{\centerline{\includegraphics[width=#1\columnwidth]{#2}}}
\newcommand{\showh}[2][.8]{\includegraphics[width=#1\columnwidth]{#2}}
\newcommand{\shows}[2][.8]{\centerline{\includegraphics[scale=#1]{#2}}}
\newcommand{\showhs}[2][.8]{\includegraphics[scale=#1]{#2}}
\newcommand{\mov}[2]{\movie[externalviewer]{{\color{blue}\small #1}}{movies/#2}}
\newcommand{\movex}[2]{\movie[externalviewer]{#1}{#2}} 
\newcommand{\movh}[3][autostart,loop]{
\movie[#1]{\showh[#2]{#3.jpg}}{#3.mp4}%
\movie[externalviewer]{$\circ$}{#3.mp4}
}
\newcommand{\movc}[3][autostart,loop]{\centerline{\movh[#1]{#2}{#3}}}
\newcommand{\mymovie}[3][]{\centerline{\movie[autostart,loop,#1]{\includegraphics[#1]{#2}}{#3}}}

\newcommand{\cen}[1]{\centerline{#1}}

\newcommand{\citing}[1]{
{\color{citcol}\ttiny#1\par}
}

\newcommand{\cit}[3]{
\par\smallskip
{\color{citcol}\ttiny #1: \emph{#2}. #3 \par}
}

\newcommand{\citurl}[4]{
\par\smallskip
{\color{citcol}\ttiny #1: \protect{\href{#4}{\color{blue}{#2.}}} #3 \par}
}

\newcommand{\cito}[3]{
\par\smallskip
{\color{bluecol}\ttiny #1: \emph{#2}. #3 \par}
}

\newcommand{\redoMacrosInProof}{
  \renewcommand{\d}{\delta}
  \renewcommand{\=}{\!=\!}
}

\newcommand{\pdflatex}{
  \definecolor{bluecol}{rgb}{0,0,.5}
  \DeclareGraphicsExtensions{.pdf,.png,.jpg,.eps}
  \renewcommand{\r}{\varrho}
  \renewcommand{\l}{\lambda}
  \renewcommand{\L}{\Lambda}
  \renewcommand{\s}{\sigma}
  \renewcommand{\b}{\beta}
  \renewcommand{\d}{\delta}
  \renewcommand{\k}{\kappa}
  \renewcommand{\t}{\theta}
  \renewcommand{\O}{\Omega}
  \renewcommand{\o}{\omega}
  \renewcommand{\SS}{{\cal S}}
  \renewcommand{\=}{\!=\!}
}

\newcommand{\mypause}{\pause}
\newcommand{\defi}[1]{\textbf{#1}}
\newcommand{\red}[1]{\emph{\color{red}#1}}
\newcommand{\pos}{{\textsf{pos}}}
\newcommand{\eff}{{\textsf{eff}}}
\newcommand{\rot}{{\textsf{rot}}}
\newcommand{\veC}{{\textsf{vec}}}
\newcommand{\quat}{{\textsf{quat}}}
\newcommand{\col}{{\textsf{col}}}
\newcommand{\de}[2]{\frac{\partial #1}{\partial #2}}
\newcommand{\target}{{\text{target}}}
\newcommand{\near}{{\text{near}}}
\newcommand{\qfree}{Q_{\text{free}}}
\renewcommand{\vec}{\boldsymbol}
\newcommand{\lft}{\text{left}}
\newcommand{\rgh}{\text{right}}
\newcommand{\prev}{{\text{prev}}}
\newcommand{\TR}[2]{T_{{#1}\to{#2}}}
\newcommand{\RO}[2]{R_{{#1}\to{#2}}}
\newcommand{\liter}{\helvetica{8}{1.1}{m}{n}\parskip 1ex}
\newcommand{\Fc}{\color{green}F}
\newcommand{\muc}{\color{blue}\mu}
\newcommand{\Astar}{A$^*$}

\providecommand{\defn}[1]{\textbf{#1}}

\newcommand{\adec}{\r_\a^-}
\newcommand{\ainc}{\r_\a^+}
\newcommand{\ldec}{\r_\l^-}
\newcommand{\linc}{\r_\l^+}
\newcommand{\minc}{\r_\m^+}
\newcommand{\mdec}{\r_\m^-}
\newcommand{\step}{{\delta}}
\newcommand{\lsstop}{\r_{\text{ls}}}
\newcommand{\stepmax}{\step_{\text{max}}}
\newcommand{\Min}{\text{min}}
\newcommand{\Max}{\text{max}}
\DeclareMathOperator*{\argmin}{arg\,min}
\DeclareMathOperator*{\EEE}{\mathbb{E}}

\definecolor{amaranth}{rgb}{0.9, 0.17, 0.31}
\definecolor{BLUE}{rgb}{0,0,1}
\definecolor{algblue}{HTML}{125DA6} 
\definecolor{linkblue}{RGB}{34,121,181}
\newcommand{\rred}[1]{\textcolor{amaranth}{#1}}
\newcommand{\bblue}[1]{\textcolor{algblue}{#1}}
\newcommand{\marc}[1]{\textcolor{green}{marc: #1}}
\newcommand{\todo}[1]{\textcolor{amaranth}{TODO: #1}}

\newcommand{\wrt}{w.r.t.\@\xspace}
\newcommand{\cma}{\Delta \vx_{\textsc{cma}}}

\begin{document}

\maketitle

\begin{abstract}
\myAbs\footnote{Project website with videos and code: \url{https://sites.google.com/view/trajectoryfirst}.}
\end{abstract}

\section{Introduction}
While reinforcement learning (RL) has achieved remarkable success in complex tasks, most RL methods assume a unimodal action distribution and produce only a single policy.
In contrast, humans and animals can solve the same task using multiple qualitatively different strategies.
A lack of behavioral diversity may lead to agents that are brittle to environmental perturbations and thereby trapped in local optima~\citep{page2017diversity, hong2004groups}, which is a core challenge to deploying robust autonomous agents. 
Therefore, this work considers the discovery of a policy \emph{set} that maximizes the reward in diverse ways.

A number of previous works have investigated this problem from various perspectives.
Notably, the fields of Novelty Search (NS) and Quality-Diversity (QD) have proposed algorithms that populate an archive of solutions based on their novelty and performance~\citep{lehman2011evolving, lehman2011novelty, conti2018improving}.
Gradient-based RL approaches instead train multiple policies in parallel on intrinsic diversity rewards that are combined with extrinsic task rewards using Lagrange multipliers~\citep{zahavy2023discovering}, bandits~\citep{parker2020effective}, or linear combinations~\citep{kumar2020one, Masood2019DiversityInducingPG, gangwani2018learning}.
While these approaches have produced impressive outcomes, they often implicitly assume that sufficiently diverse high-reward behaviors are easily reachable under the data distribution induced by online policy optimization.
As a result, most prior work focused on free-space locomotion and navigation domains~\citep{eysenbach2018diversity, zahavy2023discovering, cheng2024learning, vlastelica2024offline, g2024discovering} where different strategies tend to be easier to discover~\citep{choi2024unsupervised}.
However, when exploration is challenging, diversity objectives can become ineffective because alternative strategies are too rarely discovered to be reinforced, and the policy set may collapse to a single mode as a result.

We illustrate this issue on the example of robot manipulation.
In manipulation, object states are mediated by intermittent contacts, and qualitatively different strategies often correspond to different contact sequences.
This yields sparse diversity signals, as the mapping from robot actions to diverse object states is non-smooth and dominated by contact discontinuities~\citep{pang2023global}.
As a result, diversity-seeking objectives may still collapse to a few modes and remain under-evaluated in challenging contact-rich tasks~\citep{rho2024language, emukpere2024slim}, a finding that we corroborate in this work.

Inductive biases have driven significant advances in RL~\citep{pateria2021hierarchical, battaglia2018relational, ramesh2023physics}.
We argue that diversity optimization also benefits from inductive biases, particularly those that explicitly address exploration.
We therefore propose a new and simple \textit{trajectory-first} curriculum for learning diverse policies.
In contrast to local exploration methods such as noise-based exploration, we first explore in \emph{trajectory space} which enables to effectively jump between modes and produces diverse high-reward anchors for policy learning.
Concretely, the curriculum \textit{(i)} uses an evolutionary search over open-loop action sequences to uncover a diverse set of high-reward behaviors and \textit{(ii)} distills these behaviors into distinct, off-policy, model-free policies. 
While prior work proposed similar formulations that first solve exploration and then learning~\citep{campos2020explore, nair2018overcoming}, we do not rely on human demonstrations and propose an evolutionary approach to maximize diversity at the trajectory level instead of optimizing neural-network parameters during exploration.

\begin{figure}[t]
    \centering
    \begin{minipage}[c]{0.3\textwidth}   
    \includegraphics[width=\linewidth, page=7, trim={3cm .5cm 3cm .4cm}, clip]{imgs/rlc_plot.pdf}
    \end{minipage}
    \hspace{.1cm}
    \begin{minipage}[c]{0.6\textwidth}
        \includegraphics[width=\linewidth]{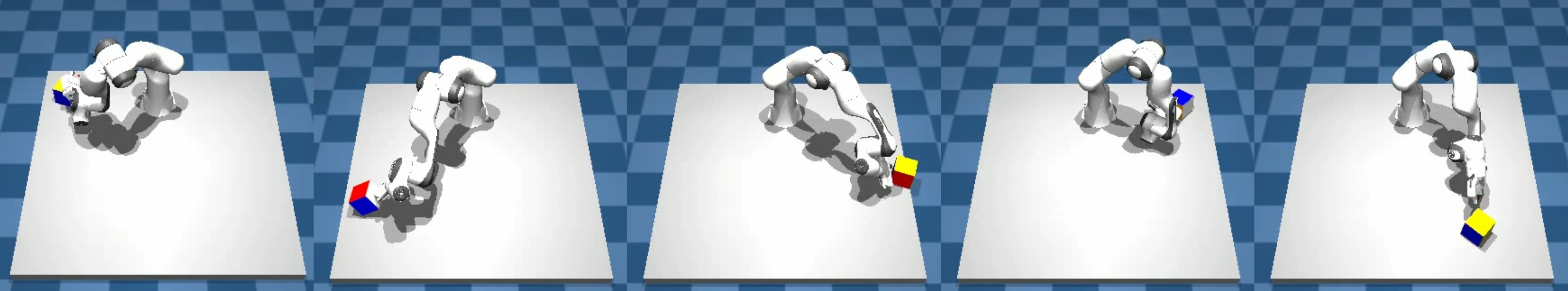}\\[0.25cm]
        \includegraphics[width=\linewidth]{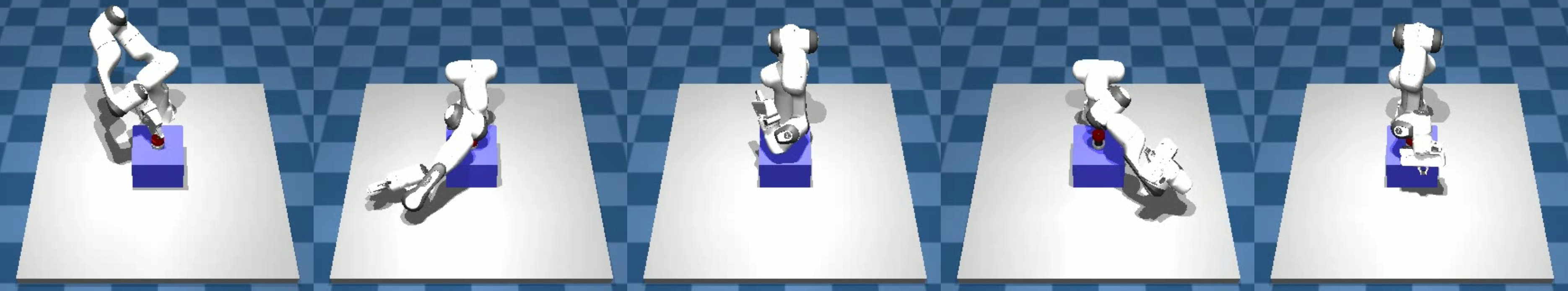}
    \end{minipage}
    
    \caption{\textbf{Left: } Overview of the proposed diversity curriculum.
    We use a spline-based trajectory prior to improve exploration.
    First, an evolution strategy explores in trajectory space to maximize diversity of trajectory parameters $\omega \in \Omega$ under performance constraints.
    Then, this data is used to warmstart the online training of multiple RL agents $\pi \in \Pi$ to solve the same optimization problem in policy space.
    \textbf{Right:} Diverse and near-optimal box pushing and button pressing policies discovered by our approach.
    }
    \label{fig:overview}
\end{figure}

In short, we make three contributions with this work. 
First, we empirically demonstrate that exploration is a key challenge, limiting existing constrained diversity optimization methods in the context of robot manipulation (\Cref{fig:rl_main}).
Second, we propose a novel curriculum for diversity optimization under extrinsic task rewards (see \Cref{fig:overview}).
Finally, we present a stabilization recipe, including a \textit{$v_{\max}$-trick} and high update-to-data ratios, to effectively distill diverse trajectories into a set of reactive policies (\Cref{sec:learning}).

\section{Preliminaries}
\textbf{Markov Decision Process.}
We model each task as a discrete-time Markov Decision Process (MDP) $M = (\SS, \AA, p, r, \gamma)$. 
At each time step $t$, the agent in state $s_t \in \SS$ selects action $a_t \in \AA$, transitions to $s_{t+1}$ with probability $p(s_{t+1} \mid s_t, a_t)$, and receives reward $r_t \triangleq r(s_t, a_t) \in [r_{\min}, r_{\max}]$. 
The objective is to learn a policy $\pi_\theta(a\mid s)$, parameterized by $\theta \in \mathbb{R}^d$, that maximizes the discounted return $J(\pi) = \EEE_{(s_t, a_t) \sim \pi} \left[ \sum_{t=\ell}^\infty \gamma^{t-\ell}r_t \right]$ with discount factor $\gamma \in [0,1)$.
We denote by $\rho_\pi(s,a)$ the state–action occupancy measure and by $\rho_\pi(s)$ its marginal over states following~\cite{haarnoja2018soft}.

\textbf{Constrained Diversity Optimization.}
While earlier works used scalars to balance diversity and task rewards, \citet{zahavy2023discovering} introduced the following constrained MDP formulation:
\begin{equation}
    \max_{\Pi^n}~ \text{Diversity}(\Pi^n)
    \quad \text{s.t.}\quad
    J(\pi) \ge \alpha v^*, \qquad \forall\, \pi \in \Pi^n
    \label{eq:domino}
\end{equation}
where $\Pi^n$ is the current set of policies, $v^* \triangleq \max_{\pi} J(\pi)$ is the value of the optimal policy in the MDP and $\alpha \in [0, 1]$ is a hyperparameter controlling the optimality constraint.
This constrained optimization problem can then be solved efficiently using dual descent on the Lagrange multipliers~\citep{altman2021constrained, borkar2005actor}.
We adopt the same problem formulation in this work.
To quantify diversity, we will measure the (squared) distance to the nearest neighbor, which shall be maximized:
\begin{equation}
    \text{Diversity}(\Pi^n) \coloneq \frac{1}{n}\sum_{i=1}^n \min_{\pi_j \neq \pi_i} \lVert \EEE_{s \sim \rho_{\pi_i}}[\phi(s)] - \EEE_{s \sim \rho_{\pi_j}}[\phi(s)] \rVert^2,\label{eq:diversity}
\end{equation}
where $\phi(\cdot): \SS \to \RRR^f$ are state-based features, which generally can be manually defined or learned, for instance using the successor feature method \citep{barreto2017successor, abbeel2004apprenticeship}.
To optimize Eq.~\eqref{eq:domino}, the common framework is to employ one-hot skill encodings $z(s_t) \in \{0,1\}^n$ as conditioning for a single policy and $Q$-function~\citep{eysenbach2018diversity, zahavy2023discovering}.
We follow this approach in our work and slightly abuse notation by writing $z(s_t)$ to denote the skill of a state and $z(\tau)$ for the skill of a full trajectory.

\textbf{Novelty Search.}
Novelty search techniques are rooted in evolutionary search.
Most novelty-search approaches to skill learning maximize diversity by using the entropy of the current policy set as an intrinsic reward \citep{conti2018improving, lehman2011evolving}.
A common choice is the particle-based $k$-NN entropy estimator~\citep{singh2003nearest}, which estimates the sparsity of the distribution based on the distance between the datapoints $\{x_i\}_{i=1}^n$ and their $k$-th nearest neighbor $\HH(X) \propto \sum_{x_i \in X} \log \lVert x_i - x_i^{(k)}\rVert$,
where $X=\{x_i\}_{i=1}^n$ are samples (or policy embeddings) and $x_i^{(k)}$ denotes the $k$-th nearest neighbor of $x_i$.
For $k=1$, the estimator $\hat{\HH}$ reduces to a sum of log nearest-neighbor distances:
\begin{equation}
\hat{\HH}(X) \coloneq \sum_{x_i \in X} \log \left( \min_{x_j \neq x_i} \lVert x_i - x_j\rVert \right),\label{eq:entropy}
\end{equation}
making its close relation to the nearest-neighbor diversity objective in Eq.~\eqref{eq:diversity} immediate when $x_i$ is chosen as a policy embedding, i.e., $x_i=\mathbb{E}_{s\sim\rho_{\pi_i}}[\phi(s)]$.

\section{Evolutionary Exploration for Diverse Policy Discovery}\label{sec:method}
To prevent diversity optimization from collapsing into local optima, we propose a two-stage curriculum: \textit{(i)} exploring the space of \emph{trajectories} via an evolution strategy (ES) to find diverse anchor behaviors, followed by \textit{(ii)} distilling them into a set of reactive \emph{policies} using stabilized constrained off-policy learning.

\subsection{Constrained Novelty Search for Spline-Based Exploration}\label{sec:cns}
The first stage of our curriculum directly optimizes agent trajectories $\tau \in \RRR^{T \times u}$, where $T$ is the horizon and $u$ the control dimensionality.
To provide a smooth inductive bias, we parameterize $\tau$ as a B-spline defined by control points $\omega \in \mathbb{R}^{m \times u}$.
Our goal is to find a trajectory set $\Omega^n = \{\omega_i\}_{i=1}^n$ that maximizes diversity subject to near-optimality:
\begin{equation}
\max_{\Omega^n} \text{Diversity}(\Omega^n) \quad \text{s.t.} \quad v_{\omega_i} \geq \alpha v^*, \quad \forall \omega_i \in \Omega^n, \label{eq:ours_cns}
\end{equation}
where $v_{\omega_i} = J^{\text{ext}}(\tau(\omega_i)) = \tsum_{t=0}^Tr^{\text{ext}}(s^i_t, a^i_t)$ is the undiscounted return and $v^*$ is the optimal task performance.
To estimate the trajectory value $v_{\omega_i}$ we roll out the interpolated controls in the MDP and collect the undiscounted return.
Since $r^{\text{ext}}$ is often non-differentiable and contact-rich, we opt for blackbox optimization to solve Eq.~\eqref{eq:ours_cns}.

Specifically, we introduce a diversity-seeking trajectory optimization approach, \textbf{Constrained Novelty Search (CNS)}, which maintains one search distribution per trajectory parameter vector $\omega_i$. 
We optimize the entropy $\hat{\mathcal{H}}$ of the set (Eq.~\eqref{eq:entropy}) via a Lagrangian objective:
\begin{equation}
    \min_{\lambda_i \geq 0} \max_{\omega_i} \mathcal{L}(\omega_i, \lambda_i) = J^{\text{int}}(\tau_i) + \lambda_i (v_{\omega_i} - \alpha v^*), \qquad \lambda_i \geq 0.
\end{equation}
To ensure stability, we follow \citet{zahavy2023discovering} and bound the multipliers using a sigmoid $\sigma(\lambda_i)$, yielding the weighted objective:
\begin{align}
    \mathcal{L}(\tau_i, \lambda_i) &= \sum_{(s^i_t, a^i_t) \in \tau_i}~[1 - \sigma(\lambda_i)]~r^{\text{int}}(s^i_t) + \sigma(\lambda_i)~r^{\text{ext}}(s^i_t, a^i_t)\nonumber\\
    &= \sum_{(s^i_t, a^i_t) \in \tau_i} \left[ [1 - \sigma(\lambda_i)]~\log \left(\min_{\tau(\omega_j) \in \Omega^n} \lVert \phi(s_t^i) - \phi(s^j_t)\rVert_2 \right) + \sigma(\lambda_i)~r^{\text{ext}}(s^i_t, a^i_t) \right],\label{eq:cns}
\end{align}
where $\phi$ is a feature extraction function that projects the states to a lower dimension.
We optimize $\omega_i$ using CMA-ES~\citep{hansen2001}, which is highly efficient for the moderately sized parameter space of B-splines ($m\times u \ll d_{\text{policy}}$). 
This is empirically more sample-efficient than using isotropic search distributions, as it was proposed in prior novelty-seeking blackbox optimization approaches~\citep[see \Cref{sec:cns_res} for further discussion on this]{conti2018improving, parker2020effective}.
The multipliers $\lambda_i$ are updated via gradient-based dual descent on the suboptimality gap.
The full procedure is outlined in more detail in \Cref{alg:ours}.

\subsection{Efficient Policy Diversity Optimization with Prior Data}\label{sec:learning}
In the second stage, we distill the diverse CNS trajectories $\mathcal{D}=\{(\tau_i, z_i)\}_{i=1}^C$ into a set $\Pi^n$ of $n$ reactive, skill-conditioned \emph{policies} $\pi_z$ that preserve diversity while satisfying the near-optimality constraint in \Cref{eq:domino}.
We adopt the Domino framework \citep{zahavy2023discovering}, treating the gradient of the diversity objective as an intrinsic reward: 
\begin{equation}
    r^{\text{int}}(s_t,a_t \mid z) = \phi(s_t)^\top \left(\bar{\phi}_z - \bar{\phi}_j\right), \quad \st \quad \bar{\phi}_j = \argmin_{j \neq z}~\lVert \bar{\phi}_z - \bar{\phi}_j \rVert^2_2,\label{eq:domino_int}
\end{equation}
where $\bar{\phi}_z = \EEE_{s \sim \rho_{\pi_z}}[\phi(s)]$ are the expected features per skill.
The resulting policy objective maximizes the expected cumulative discounted extrinsic and intrinsic rewards:
\begin{equation}
    \min_{\lambda_i \geq 0}\max_{\pi_i}~\EEE_{s_0, \pi_i} \left[ \sum_{t=1}^T\gamma^t\left( [1 - \sigma(\lambda_i)]~r^{\text{int}}(\pi_{\theta_i}) + \sigma(\lambda_i)~r^{\text{ext}}(\pi_{\theta_i}))\right) \right].\label{eq:domino_full}
\end{equation}
As in the CNS step, we alternate policy and multiplier updates.
The multipliers are updated on the suboptimality gap $\min_{\lambda_i} \lambda_i(v_{\pi_i} - \alpha v^*)$.
We follow Domino in estimating policy values and features using their undiscounted average over time, i.e., $v_i = \lim_{T \to \infty} \tfrac{1}{T} \EEE_{s_0, \pi_i}\tsum_{t=1}^T r_t$ and $\bar{\phi}_i = \lim_{T \to \infty} \tfrac{1}{T}\EEE_{s_0, \pi_i}\tsum_{t=1}^T \phi(s_t)$.

Although the Domino reward in Eq.~\eqref{eq:domino_full} permits balancing task and diversity objectives, distilling reactive policies from prior data under this reward is nontrivial.
To ensure stable distillation in contact-rich tasks, we introduce three critical modifications to standard constrained learning:

\textbf{1. Symmetric Sampling.} To prevent diversity collapse during the transition to online RL, we use a stratified buffer and \textit{symmetric sampling}~\citep{ball2023efficient, vecerik2017leveraging, ross2012agnostic}.
Each training batch is composed of 50\% online data and 50\% filtered CNS trajectories. 
The latter are selected from the top 25\% of samples based on their returns.
This ensures efficient learning that does not suffer from diversity collapse as the policies become reactive.
In addition, we initialize the expected features for Eq.~\eqref{eq:domino_int} with the averages over the CNS dataset.

\textbf{2. The $\mathbf{v_{\max}}$-trick.} 
Prior work noticed that the optimization of \Cref{eq:domino} can be unstable and challenging~\citep{fu2023iteratively}.
We identify the estimation of the optimal value $v^*$ as key for learning stabilization.
Moving it too quickly will suppress diversity, and induce high target non-stationarity while moving it too slowly prevents efficient learning.
While \citet{zahavy2023discovering} train an expert policy that only maximizes extrinsic reward to estimate $v^* = v_{\pi_{\text{ext}}}$, we propose
\begin{equation}
    v^* \leftarrow \max \{v^*, \max_i v_i\}, \label{eq:vmax}
\end{equation}
which tracks the maximum performance achieved across all skills (including a dedicated extrinsic-only expert) and throughout the training history. 
This provides a monotonically increasing, stable target for the suboptimality constraint.
To prevent underestimation, we still train a dedicated expert policy but reduce the fluctuations of value estimates via taking the historical maximum over the entire policy set.

\textbf{3. High Policy UTD.} 
Standard offline-to-online RL often uses a high update-to-data (UTD) ratio for the critic while keeping the policy update infrequent~\citep{ball2023efficient}. 
The intuition is to quickly fit the critic on the guiding offline data without overfitting the policy to recent online experience.
However, in diversity optimization, the intrinsic reward is inherently non-stationary as it depends on the state occupancy of the current policy set. 
We therefore propose to reduce the update asymmetry between actor and critic for diversity-seeking offline-to-online RL.
Updating the policy multiple times per environment step is essential to keep pace with the shifting intrinsic landscape, leading to significantly faster and more stable mode discovery.

\section{Related Work}
\textbf{Diversity-Driven Policy Discovery.}
Behavioral diversity objectives in policy learning have been approached with techniques from two main categories, namely gradient-free policy search and classical gradient-based RL.
Quality-Diversity (QD) and evolutionary methods search in a gradient-free manner, populating archives of high-performing, behaviorally distinct solutions~\citep{mouret2015illuminating, cully2015robots} or co-optimizing fitness and novelty across populations~\citep{ulrich2011maximizing, parker2020effective, braun2025stein}.
In principle, these approaches could be used in the first step of the proposed curriculum.
However, most prior work focuses on optimizing in policy parameter space before distilling policies \citep{faldor2023map, mace2023quality, batra2024proximal}, a less effective process as we find in this work.
In gradient-based RL, diversity is often tackled via intrinsic motivation bonuses~\citep{eysenbach2018diversity, Sharma2020Dynamics-Aware, fu2023iteratively, chen2024dgpo, celik2024acquiring}. 
Recent advances have shifted the focus from intrinsic reward definition toward optimization frameworks. 
For instance, \citet{celik2024acquiring} proposed a mixture-of-experts approach for training policies that specialize on different conditions in the environment. 
Concurrently, Lagrange multipliers have emerged as a key technique for balancing task optimality against behavioral diversity~\citep{zahavy2023discovering, grillotti2024quality, vlastelica2024offline, g2024discovering}.
While most of these works focus on the optimization perspective of diverse policy learning, i.e., when to update which policy and how, they often assume that a diversity objective alone can drive discovery.
However, we show that this is insufficient for discovering globally distinct modes in complex robot manipulation tasks.

\textbf{Exploration in RL.} 
Exploration is a fundamental aspect of RL enabling agents to effectively learn and generalize. 
Effective exploration in robotics remains a fundamental challenge due to the sparsity of high-reward regions and the complexity of contact dynamics.
Common strategies include local noise-based exploration~\citep{haarnoja2018soft, fortunato2018noisy}, knowledge-based exploration~\citep{burda2019rnd}, and empowerment-based exploration~\citep{houthooft2016vime, eysenbach2018diversity, laskin2022cic}.
However, these typically operate at the level of step-wise action selection. 
To address this, recent work has explored trajectory-level learning, often using movement primitives as action representations~\citep{otto2023deep, klink2020self}.
While effective for discovery, open-loop action sequences lack the reactivity required to handle perturbations and error recovery.
Our work bridges this gap as we leverage the exploration benefits of trajectory-level search via B-splines, but unlike prior episodic methods~\citep{celik2024acquiring}, we distill these behaviors into reactive, closed-loop policies.

\textbf{RL Finetuning from Offline Data.} 
The second stage of our curriculum relates to the field of offline-to-online RL, where agents have access to a static offline dataset as well as the environment for learning~\citep{zhou2025efficient}.
Key challenges in this regime are potential distributional shifts and catastrophic forgetting of the offline prior during online adaptation~\citep{chen2025offline}. 
Strategies such as data retention and symmetric sampling have proven effective for maintaining task performance~\citep{ball2023efficient, nair2018overcoming}. 
Crucially, however, prior work focuses almost exclusively on maximizing a \emph{single} optimal policy. 
We extend this paradigm to the multi-policy setting, investigating how data selection and sampling strategies influence the retention and amplification of diversity.


\section{Experiments}\label{sec:exp}
\begin{figure}[t]
    \centering
    \includegraphics[width=.22\linewidth, trim={1cm 0cm 1cm 0.2cm}, clip]{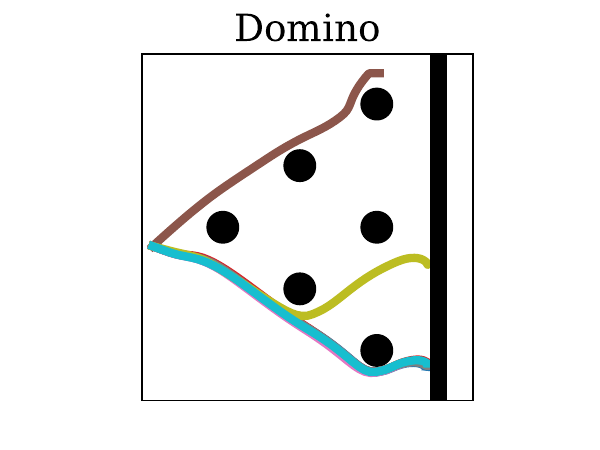}
    \includegraphics[width=.22\linewidth, trim={1cm 0cm 1cm 0.2cm}, clip]{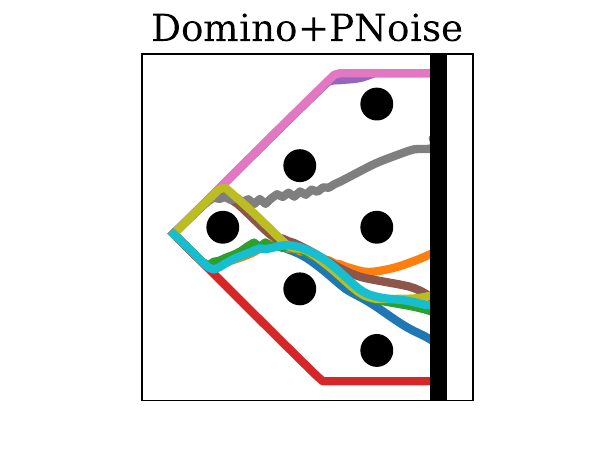}
    \includegraphics[width=.22\linewidth, trim={1cm 0cm 1cm 0.2cm}, clip]{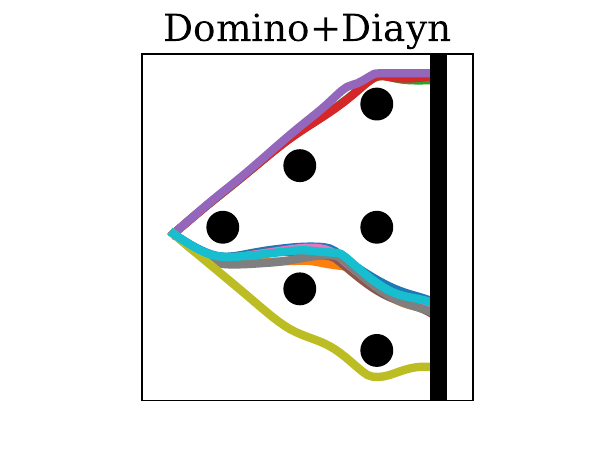}
    \includegraphics[width=.22\linewidth, trim={1cm 0cm 1cm 0.2cm}, clip]{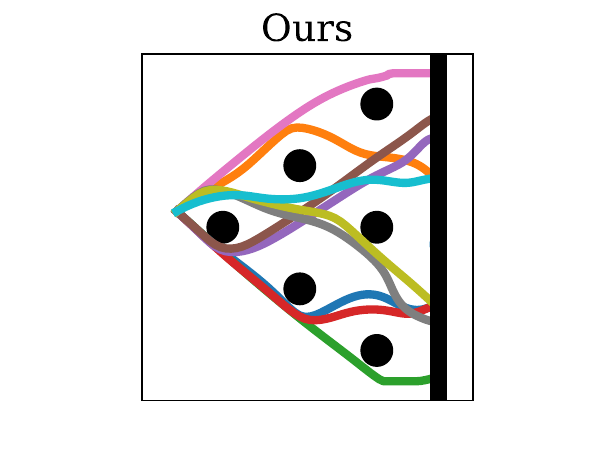}
    \caption{Final trajectories on the maze navigation tasks. Each color depicts a different skill $z_i$. Standard constrained diversity optimization (Domino) underexplores the environment. We see that our method discovers the highest number of distinct paths through the maze.}
    \label{fig:pointmass}
\end{figure}

Our evaluation characterizes the limitations of contemporary diversity-seeking RL and demonstrates the efficacy of trajectory-level curricula. 
Specifically, we investigate:
\begin{enumerate}[label=(\roman*)]
    \item \textbf{Exploration Failure:} Does local policy-space exploration in constrained diversity optimization (e.g., Domino) discover diverse policies in robot manipulation tasks?
    \item \textbf{Curriculum Effectiveness:} Can a trajectory-level prior effectively anchor agents in diverse, high-reward functional modes?
    \item \textbf{Stabilization Analysis:} Which algorithm components are necessary to prevent diversity decay during the transition from open-loop trajectories to reactive policies?
    \item \textbf{Reward Robustness:} How robust is our method to reward landscapes characterized by high penalties, as they are common in robotics?
\end{enumerate}

\begin{wrapfigure}{R}{0.6\textwidth}
  \begin{center}
    \includegraphics[width=0.29\textwidth, trim={0.25cm 0.4cm 0.25cm 0.4cm}, clip]{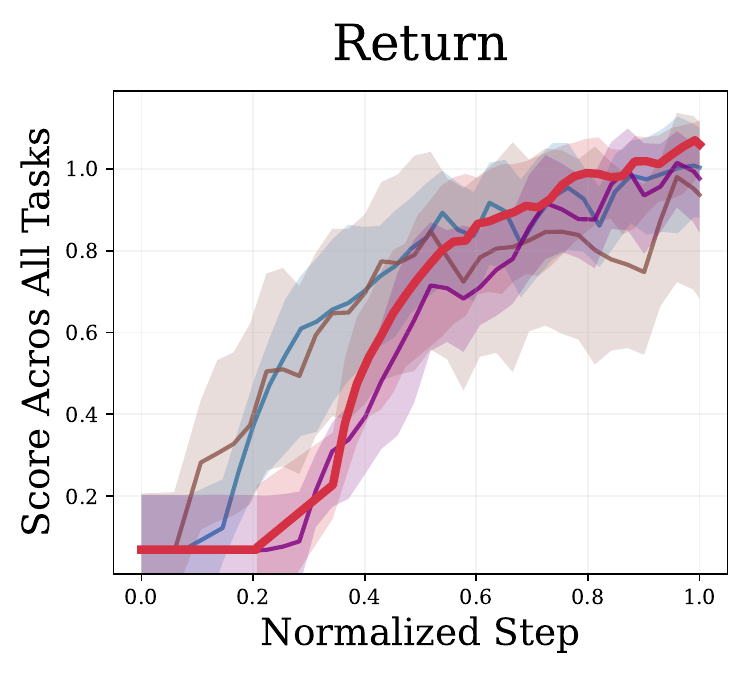}
    \includegraphics[width=0.29\textwidth, trim={0.25cm 0.4cm 0.25cm 0.4cm}, clip]{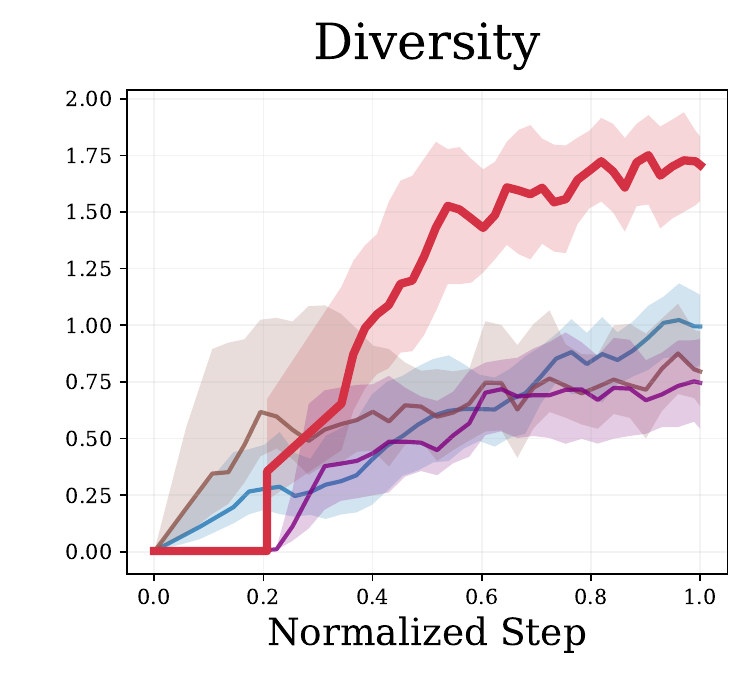}
  \end{center}
  {\hspace{.2cm} \includegraphics[width=.95\linewidth, trim={0.2cm 0.5cm 0.2cm 0.5cm}, clip]{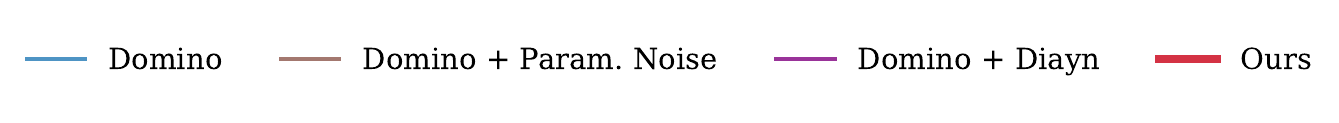}}\\[-.6cm]
  \caption{Results of benchmarking the proposed curriculum exploration against other diversity-seeking RL methods. We report the normalized performances across all tasks relative to plain Domino using IQM and $\pm95\%$-CI computed over 5 seeds. Individual per-task learning curves are listed in \Cref{sec:learn_curves}.}
  \label{fig:main_res}
\end{wrapfigure}

\subsection{Experimental Setup}\label{sec:setup}
\textbf{Task Environments.} We evaluate on four robotic tasks designed to test diversity across navigation and robotic manipulation. 
All environments are depicted in Fig.\ \ref{fig:envs} in the appendix.
(1) \textbf{Maze Navigation:} A robot must navigate a rod attached to its gripper through a maze without collisions. 
Diversity is measured on the rod trajectories in the $xy$-plane across 10 skills. 
(2) \textbf{Cube Pushing:} The agent must push a cube on a table as far as possible. 
Diversity is defined by the object trajectories in the $xy$-plane across 5 skills. 
(3) \textbf{Button Press:} The robot must press a button. 
Diversity is measured via contact forces between the button and different robot links as well as robot poses, encouraging whole-body manipulation strategies. 
We train 5 skills on this task.
(4) \textbf{Cube Flip:} A task requiring the agent to flip a cube, possibly using environmental structures. 
This task assesses the performance of our method in contact-rich tasks that require precise manipulation.
Diversity is measured using the contact forces between the cube and the environmental structure to quantify the contact mode variation.
We again train 5 skills on this task.
For full environment details, we refer to \Cref{sec:exp_dets}.
We encourage viewing the supplementary video that displays our policies and contrasts them with prior methods.

\textbf{Setting and Baselines.} 
To guarantee a fair comparison, all baselines are allocated the same total number of environment steps for online training as our curriculum uses for CNS and learning together.
We compare against: (i) \textbf{Domino} \citep{zahavy2023discovering}, representing state-of-the-art constrained diversity optimization; (ii) \textbf{Parameter Noise} \citep{fortunato2018noisy} for policy parameter-space exploration; and (iii) \textbf{DIAYN} \citep{eysenbach2018diversity} as an unsupervised skill discovery baseline, which we fine-tune with the Domino objective to maximize task performance.
All code uses SAC as the core algorithm \citep{haarnoja2018soft} and uses return normalization, critic ensembling, and layer norm.
For further implementation details, we refer to~\Cref{sec:exp_dets}.
In our evaluation we report the expected return across all policies as well as the policy diversity defined by Eq.~\eqref{eq:diversity}.
All experiments are repeated across five seeds over which we report the interquartile mean (IQM) and bootstrapped 95\% confidence intervals (CIs).

\begin{figure}[t]
    \centering
    {\hspace{.2cm} \includegraphics[width=.8\linewidth, trim={0.2cm 0.5cm 0.2cm 0.75cm}, clip]{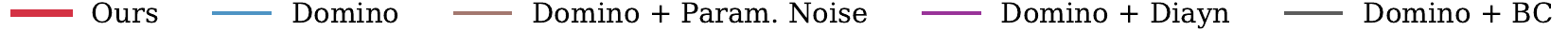}}\\[-.4cm]
    \subfloat[Maze Navigation]{\includegraphics[width=0.23\linewidth]{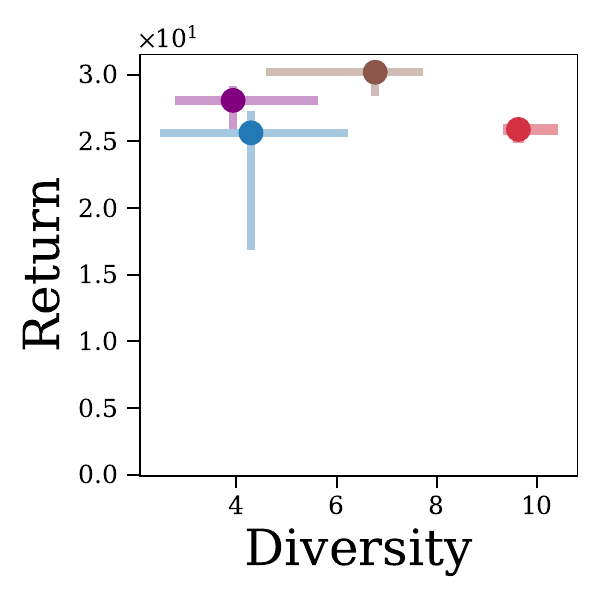}} \hfill
    \subfloat[Button Press]{\includegraphics[width=0.23\linewidth]{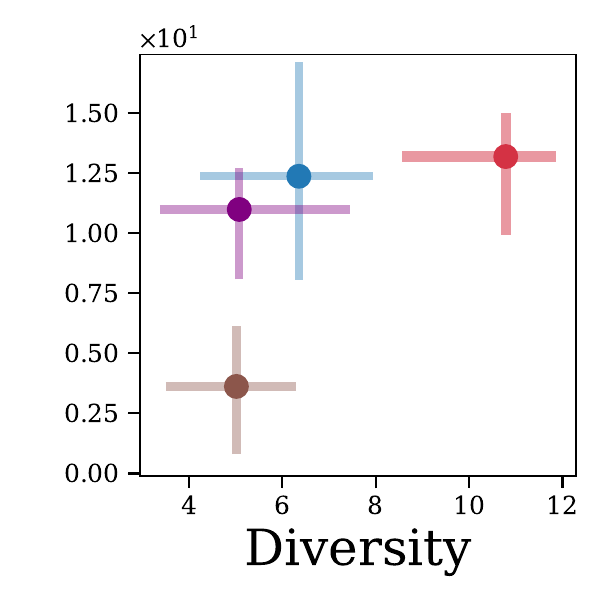}} \hfill
    \subfloat[Cube Push]{\includegraphics[width=0.23\linewidth]{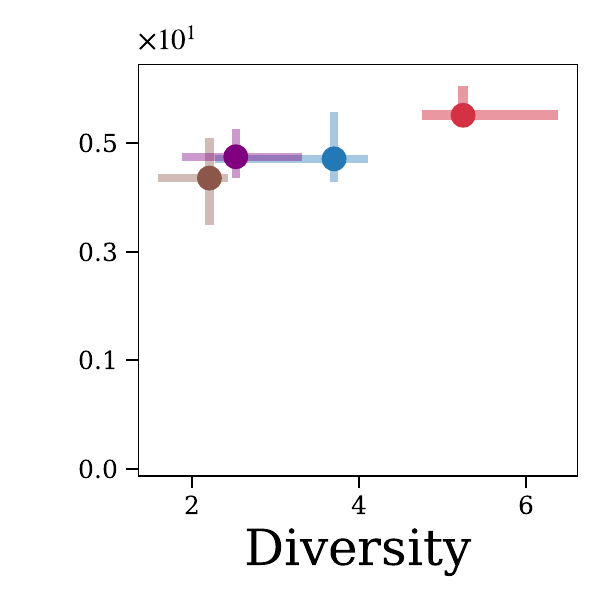}} \hfill
    \subfloat[Cube Flip]{\includegraphics[width=0.23\linewidth]{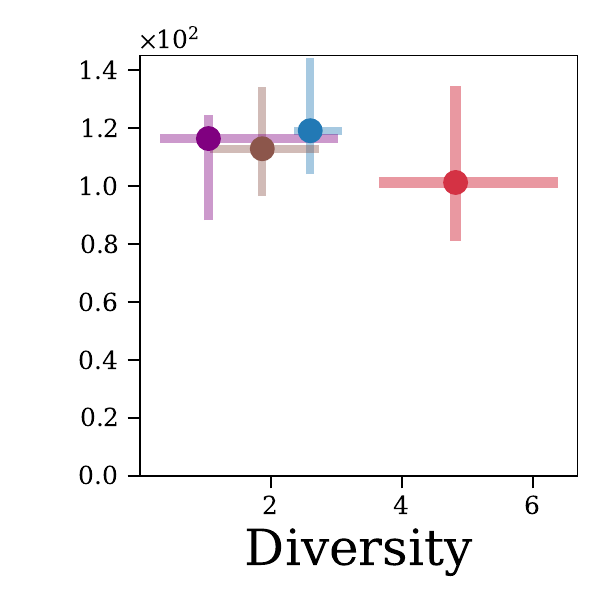}} 
    \caption{Final results across the four tasks. Our method finds more diverse solutions than the baselines without compromising on task performance. We report the IQM of final returns and diversity across 5 seeds as well as $95\%$ bootstrapped CIs.}
    \label{fig:rl_main}
\end{figure}

\subsection{Analysis of Constrained Policy Optimization Bottlenecks}\label{sec:shortcomings}
Investigating the performance of default constrained diversity optimization, i.e., Domino, we see that the policy sets lack diversity.
Qualitatively, we observe in Fig.~\ref{fig:pointmass} that most policies take the safe routes around the obstacle region to avoid collision, resulting in few behavioral modes being covered.
In contrast to free-space navigation, the collision penalties in this task make the behavioral modes disconnected and challenging to bridge, causing Domino to produce limited policy diversity.
We corroborate these findings in the more challenging manipulation tasks. 
While task performance is high overall, policy diversity is lacking (Fig.\ \ref{fig:main_res} \& \ref{fig:rl_main}).
For instance, in the button press task, often all policies use the same part of the robot morphology to press the button.
While the specific robot link varies across seeds, it is consistent within a single seed, which underlines that the diversity optimization only occurs within a single local optimum (Fig.\ \ref{fig:qual_res} illustrates policy snapshots).
This indicates that while diversity constraints are theoretically appealing, local gradients in policy space are insufficient to jump between different (disconnected) regions on high-reward manifolds.

\subsection{Improvements via Trajectory-Level Priors}\label{sec:contrib}
\Cref{fig:main_res} displays the task and diversity scores of all methods across all tasks. 
Our method achieves a 1.5x increase in diversity over the strongest baseline at similar task performance.
We want to highlight that this is despite the learning delay due to the CNS environment steps, as indicated by the inflection points in the per-task curves (Fig.\ \ref{fig:learn_curves}).
Specifically, we observe that the final diversity of the policy sets produced by our method is significantly higher (Fig.\ \ref{fig:rl_main}), while the speed of learning matches the baselines (Fig.\ \ref{fig:main_res}).
Despite our finding that noise-based exploration also improves over Domino in the relatively simple navigation task (\Cref{fig:pointmass}), these improvements via the baseline exploration methods do not materialize on the more challenging manipulation tasks.
Particularly on the button press task, parameter noise performs poorly, mainly because the noise-based exploration leads to frequent collisions with the button-box, resulting in negative rewards, so the final policies tend to refrain from manipulating (see Fig.\ \ref{fig:qual_res}).
Similarly, we observe that the skill-based DIAYN policy pretraining fails to increase downstream manipulation policy diversity. 
We find that this happens because the discriminator is prone to exploiting proprioceptive variances that do not change the object state, e.g., by varying joint angles in free space.
This is in line with prior work on unsupervised skill discovery in robotics, which reported little success of DIAYN for loco-manipulation~\citep{cathomen2025divide}.
In contrast, by performing CNS in the structured space of B-splines, our curriculum discovers diverse contact modes that are subsequently distilled into reactive policies.
For an in-depth discussion of this, we refer readers to \Cref{secDivMan}.

\begin{figure}[t]
    \centering
    \subfloat[$v^*$ Estimation]{%
      \begin{minipage}[t]{0.2\linewidth}\vspace{0pt}\centering
        \includegraphics[height=2.2cm]{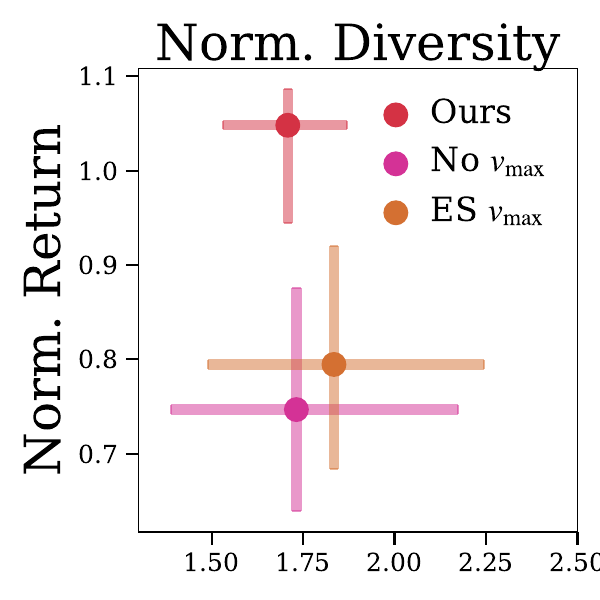}\label{fig:nomax}
      \end{minipage}
    }\hfill
    \subfloat[Symm.\ Sampling Ablation]{%
      \begin{minipage}[t]{0.25\linewidth}\vspace{0pt}\centering
        \includegraphics[height=2.2cm]{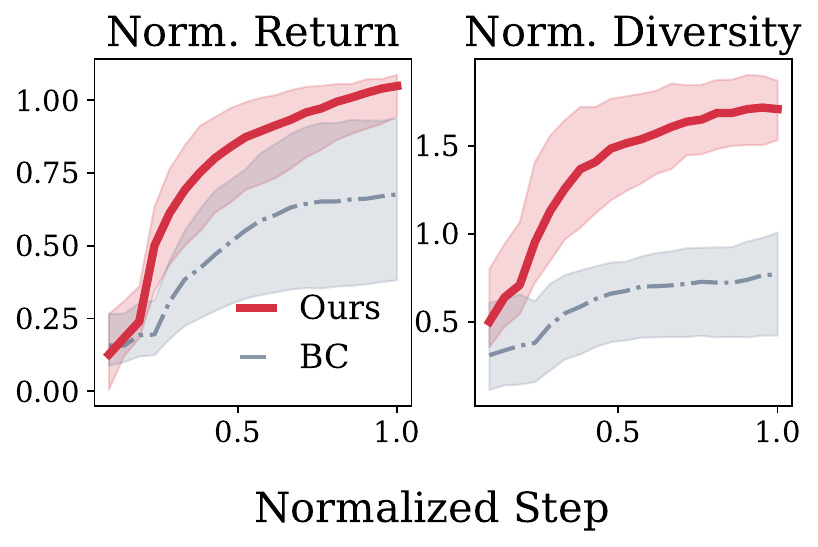}\label{fig:bcpre}
      \end{minipage}
    }\hfill
    \subfloat[Diversity Objective Ablation]{%
      \begin{minipage}[t]{0.25\linewidth}\vspace{0pt}\centering
        \includegraphics[height=2.2cm]{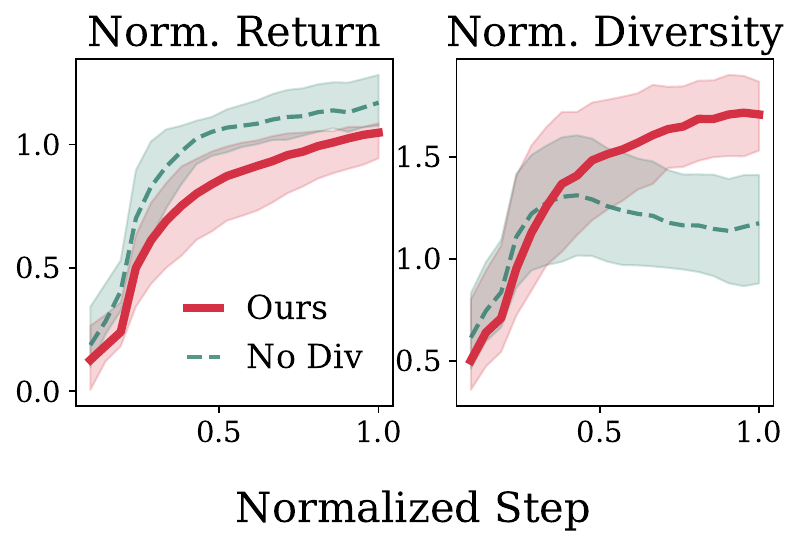}\label{fig:nodiv}
      \end{minipage}
    }\hfill
    \subfloat[Policy Delay Analysis]{%
      \begin{minipage}[t]{0.25\linewidth}\vspace{0pt}\centering
        \includegraphics[height=2.2cm]{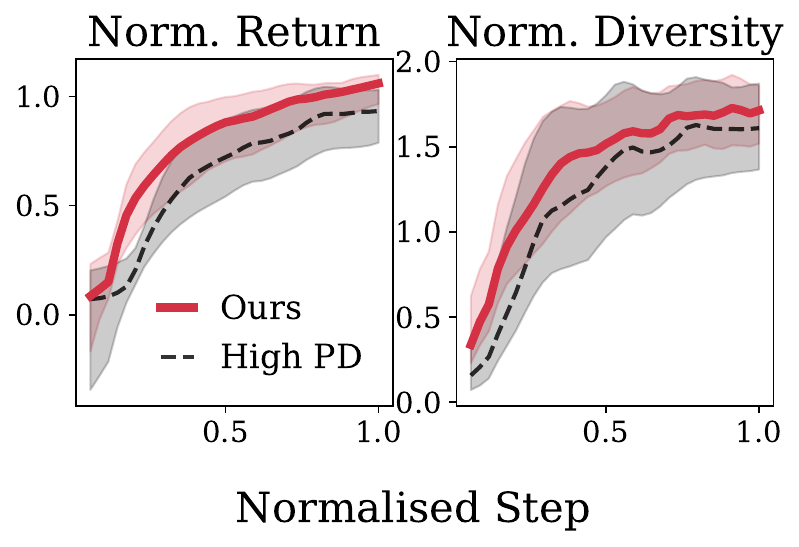}\label{fig:pidelay}
      \end{minipage}
    }
    \caption{Ablation studies: (a) When estimating the optimal MDP value $v^*$, the $v_{\max}$-\emph{trick} improves learning. (b) Symmetric sampling beats BC pretraining to incorporate CNS data. (c) Only using the diverse offline data does not suffice to learn diverse policies, a dedicated diversity objective is still required. (d) Achieving a high UTD only by updating the critic frequently is not optimal in the constrained diversity optimization setting, as the critic depends on the state occupancy under the current policy.}
    \label{fig:ablations}
\end{figure}

\subsection{Which Algorithmic Design Choices Matter?}
While we have established above that our method improves over prior work, we now discuss which aspects specifically contribute to its success.
In Fig.\ \ref{fig:ablations}, we report ablations of the key components of our learning approach aggregated across all tasks.

\textbf{Diversity optimization beats pure distillation.}
Fig.\ \ref{fig:bcpre} shows that combining Domino with symmetric sampling yields much higher returns and policy diversities than behavior cloning (BC) pretraining followed by standard Domino.
This corroborates the findings of \citet{ball2023efficient} and extends them to diversity objectives that are beyond standard linear RL objectives.

\textbf{Symmetric sampling beats BC pretraining.}
We observe in Fig.\ \ref{fig:nodiv} that simply finetuning multiple policies from the CNS data without an additional diversity objective leads to diversity collapse.
This demonstrates the importance of the diversity-focused training that we propose.

\textbf{$\mathbf{v^*}$-estimation is crucial.}
Fig.\ \ref{fig:nomax} shows that removing the $v_{\max}$-\emph{trick} yields lower return and diversity.
Interestingly, the same figure shows that initializing $v^*$ with the maximum trajectory value from the CNS data reduces performance as well.
This illustrates a key factor in constrained diversity optimization: by gradually increasing $v^*$, the diversity objective is optimized early on and not suffocated by the task objective at early stages.
This induces an implicit curriculum that gradually increases optimality pressure as $v^*$ increases over time.
Our results show that stabilizing this curriculum is key, since it reduces target variance for policies and intrinsic critics.
We further find that these findings are applicable beyond our curriculum in \Cref{fig:vmax} in the appendix, which shows improvements for plain Domino too.
Hence, we hope that they help mitigating stability-related issues that were reported related to Domino in the past~\citep{fu2023iteratively}.

\textbf{Reducing the policy-critic update asymmetry improves learning.} 
Contrary to common practices in offline-to-online RL, we find that an increased policy UTD is crucial for learning performance (Fig.\ \ref{fig:pidelay}). 
When using an UTD of 8 policy updates per (batched) environment step and Lagrange multiplier update, the performance on the cube task is significantly higher compared to only updating it once.
Commonly, it is recommended to use a high UTD for the critic to quickly learn from the offline data while learning the policy at a slower pace to not overfit to recent experience.
In our setting, however, the (intrinsic) critic depends on the state visitation distribution of the policy, which in turn depends on the Lagrange multipliers and the critic.
In this context, limiting the asymmetry of policy and critic updates stabilizes policy targets and improves overall learning performance.

\begin{wrapfigure}{R}{0.38\textwidth}
  \centering
  \begin{minipage}{0.63\linewidth}
    \centering
    \includegraphics[width=\linewidth, trim={0.25cm 0.4cm 0.25cm 0.38cm}, clip]{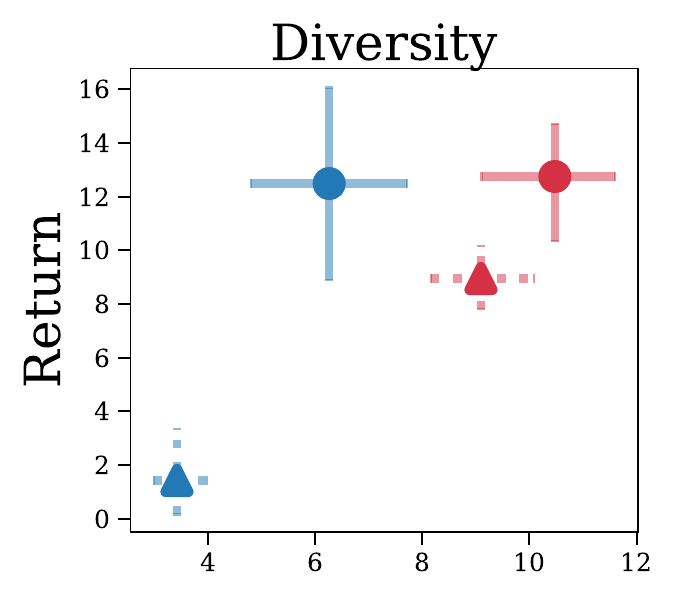}
  \end{minipage}\hfill
  \begin{minipage}{0.30\linewidth}
    \centering
    \raisebox{6mm}{%
      \includegraphics[width=\linewidth, trim={0cm 0cm 3.cm 0cm}, clip]{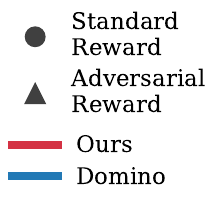}%
    }
  \end{minipage}
  \caption{Robustness analysis. Our method is more robust to reward formulation details than plain Domino.}
  \label{fig:robustness}
\end{wrapfigure}

\subsection{Robustness to Reward Specification}

In real-world robotics, reward shaping is often key to success, and common gym environments include highly engineered reward functions.
In particular, collision penalties are required for learning safe behaviors, but tuning their scale can be tedious~\citep{taomaniskill3, zakka2025mujoco}.
If their weight is too low, the robot learns unsafe policies.
Yet, if they are too high, the agent avoids all contacts so that it does not collide.
We simulate such a misspecified reward by significantly increasing collision penalties in the button press task.
Under these conditions, the exploration-exploitation trade-off in Domino breaks down, resulting in risk-averse policies that fail to interact with the object entirely.
This is reflected in final returns near zero in \Cref{fig:robustness}.
Our method remains robust because the CNS phase provides a warmstart within the high-reward region, effectively bypassing the local optima of the high-penalty landscape.
While the return reduces too, the overall return is much higher and all policies solve the task.

\section{Conclusion}\label{sec:conc}
\paragraph{Summary.}
In this work, we addressed the fundamental challenge of discovering diverse behaviors. 
Focusing on robot manipulation tasks, we identified that a major bottleneck in constrained diversity-optimizing RL is not the learning objective itself, but rather the reliance on local exploration, which frequently fails to discover diverse contact modes or paths around obstacles. 
To mitigate this, we introduced \emph{Trajectory‐First}, a two-stage curriculum that bridges global trajectory-level exploration and reactive policy control.
By anchoring the learning process in a diverse set of B-spline trajectories, we provide an inductive bias that enables the discovery of diverse policies.
Our evaluation demonstrates that our curriculum achieves a significant increase in behavioral diversity without compromising task performance. 
Furthermore, we provided a recipe for stable offline-to-online policy learning with diversity objectives, including the $v_{\max}$-trick and high policy UTD ratios. 
We believe these findings can be instrumental for future work in constrained diversity optimization.

\paragraph{Limitations and Future Work.}
Despite its effectiveness, our approach has limitations that present avenues for future research. 
First, our diversity objective relies on a state-feature mapping $\phi(s)$. 
While successor features of states can be used, we define features on rolling means of subsets of raw observations, following \citet{zahavy2023discovering}.
While this provides control over what diversity should be optimized, future work could investigate coupling our curriculum with representation learning to discover diversity directly from raw observations.
Second, our two-stage curriculum uses a fixed transition from trajectory search to policy distillation.
An adaptive switching mechanism that iterates between CNS and policy learning could further improve sample efficiency and generalization.
Finally, while we successfully demonstrated diverse manipulation policies, we focus on tasks that require little long-horizon reasoning in this work. 
Future work could extend this to discover diversity at a high-level planning and low-level control level jointly.

\section*{Acknowledgements}
This research was funded by the Amazon Fulfillment Technologies and Robotics team, as well as the German Federal Ministry of Research, Technology and Space (BMFTR) under the Robotics Institute Germany (RIG).
The authors want to thank anonymous reviewers for constructive feedback which helped improve the paper.

\bibliography{references}
\bibliographystyle{rlj}

\beginSupplementaryMaterials
\renewcommand{\thesection}{\Alph{section}}
\setcounter{section}{0}
\section{Algorithm}\label{sec:alg}
\renewcommand{\algorithmicindent}{.75em}  
\begin{algorithm}[!ht]
    \caption{Curriculum for discovering diverse policies.}
    \label{alg:ours}

    \textbf{Input:} Environment $env$, Optimality ratio $\alpha$, Num.\ skills $n$,
    CNS learning rates $\kappa^{cns}_\lambda,\kappa^{cns}_\phi,\kappa^{cns}_v$,
    RL learning rates $\kappa^{rl}_\lambda,\kappa_\xi,\kappa_V,\kappa_\pi$,
    Init.\ SAC temperature $\xi$
    
    \begin{algorithmic}[1]
        \State \textbf{// 1. Constrained Novelty Search}
        \State Initialize population parameters $\omega_i$ for skills $i = 1, \dots, n$
        \State Initialize $v^* \gets 0$, EMAs $\bar{\phi}_i, v_i$ and Lagrange multipliers $\lambda^{cns}_i \gets 0$ for $i=1,\dots,n$
        \State $\DD_{cns} \gets \{\}$

        \For{Iteration $t = 1, \dots, T$}
            \For{Population $i = 1 \dots n$}
                \State $\{\tau^i_1, \dots, \tau^i_m\} \gets \texttt{env.rollout}(\omega_i)$
                \State Update $\omega_i$ given
                $(1 - \sigma(\lambda^{cns}_i))~r^{\text{int}}(\{\tau^i_1, \dots, \tau^i_m\})
                + \sigma(\lambda^{cns}_i)~r^{\text{ext}}(\{\tau^i_1, \dots, \tau^i_m\})$ \hfill (Eq.\ \ref{eq:cns})

                \State $\bar{\phi}_i \gets (1-\kappa^{cns}_\phi)\,\bar{\phi}_i + \kappa^{cns}_\phi\,\EEE[\phi(\tau^i)]$
                \hfill (Population feature EMA)

                \State $v_i \gets (1-\kappa^{cns}_v)\,v_i + \kappa^{cns}_v\,\EEE[r^{ext}(\{\tau^i_1, \dots, \tau^i_m\})]$
                \hfill (Population value EMA)

                \State $v^* \gets \max\left\{v^*, \max \left\{v_j : j = 1,\dots,n \right\} \right\}$
                \hfill (Update estimate of $v^*$)

                \If{$t \% LAMBDADELAY = 0$}
                    \State $\lambda^{cns}_i \gets \lambda^{cns}_i - \kappa^{cns}_\lambda\,(v_i - \alpha v^*)$
                    \hfill (Lagrange loss)
                \EndIf

                \State $\DD_{cns} \gets \DD_{cns} \cup \{\tau^i_1, \dots, \tau^i_m\}$
                \hfill (Add data to buffer)
            \EndFor
        \EndFor
        \State 
        
        \State \textbf{// 2. Constrained RL Diversity Optimization}
        \State $\DD_{RL} \gets \{\}$; Filter $\DD_{cns}$ and keep top 25\% of trajectories per skill
        \State Reset $v_i \gets 0$ and $\bar{\phi}_i \gets \EEE_{s^t_i \sim \tau^i}[s^t_i]$ for $i=1,\dots,n$; set $v^* \gets 0$
        \State Initialize shared critic nets $\theta^{ext}_V, \theta^{int}_V$, policy net $\theta_\pi$ and RL multipliers $\lambda^{rl}_i \gets 0$ for $i=1,\dots,n$

        \For{Iteration $t = 1, \dots, I$}
              \State \textbf{// Environment steps (sample skill once per episode)}
              \State If new episode sample $z\sim p(z)$ (fixed for episode) 
              \State $a_t \sim \pi(a_t\mid s_t,z)$; $s_{t+1} \sim p(s_{t+1}\mid s_t, a_t)$ \hfill (Step environment)

              \State $\DD_{RL} \gets \DD_{RL} \cup \{(s_t,a_t,r^{ext}, r^{int},\phi(s_t,a_t),s_{t+1},z)\}$

              \State \textbf{// Training steps}
              \State $\mathcal{B} \gets \mathcal{B}_1 \cup \mathcal{B}_2$ with $\mathcal{B}_1 \sim \DD_{cns}$, $\mathcal{B}_2 \sim \DD_{RL}$
              \hfill (Symmetric sampling)

              \State $\theta^{ext}_V \gets \theta^{ext}_V - \kappa_V \nabla_{\theta^{ext}_V} J^{ext}_V(\theta^{ext}_V;\mathcal{B})$
              \hfill ($J^{ext}_V = \EEE_{\mathcal{B}}[\mathrm{TD}^{ext}(s)^2]$)
              \State $\theta^{int}_V \gets \theta^{int}_V - \kappa_V \nabla_{\theta^{int}_V} J^{int}_V(\theta^{int}_V;\mathcal{B})$
              \hfill ($J^{int}_V = \EEE_{\mathcal{B}}[\mathrm{TD}^{int}(s)^2]$)
               
              \State $Q^{mix} \gets [1-\sigma(\lambda^{rl}_i)]\,\hat Q^{int}(x;\theta^{int}_V) + \sigma(\lambda^{rl}_i)\,\hat Q^{ext}(x;\theta^{ext}_V)$
              \hfill (Policy target; Eq.~\ref{eq:domino_full})
              \State $J_\pi(\theta_\pi;\mathcal{B}) \gets \EEE_{(s,a)\sim\mathcal{B}}\Big[
              \xi~\log \pi_{\theta_\pi}(a\mid s) - Q^{mix}
              \Big]$
              \State $\theta_\pi \gets \theta_\pi - \kappa_\pi \nabla_{\theta_\pi} J_\pi(\theta_\pi;\mathcal{B})$
              \hfill (Update policy)

              \State $\lambda^{rl}_i \gets \lambda^{rl}_i - \kappa^{rl}_\lambda\,(v_i - \alpha v^*)$
              \hfill (Lagrange loss)

              \State \textbf{// Feature and value Update}
              \State $\bar{\phi}_i \gets (1-\kappa^{rl}_\phi)\,\bar{\phi}_i + \kappa^{rl}_\phi\,\EEE[\phi(s)]$
              \hfill (Population feature EMA)

              \State $v_i \gets (1-\kappa^{rl}_v)\,v_i + \kappa^{rl}_v\,\EEE[r^{ext}]$
              \hfill (Population value EMA)

              \State $v^* \gets \max\left\{v^*, \max \left\{v_j : j = 1,\dots,n \right\} \right\}$
              \hfill (Update estimate of $v^*$; Eq.~\ref{eq:vmax})
        \EndFor
    \end{algorithmic}
\end{algorithm}

\section{Additional Results}

\subsection{Full Learning Curves}\label{sec:learn_curves}
\begin{figure}[t]
    \centering
    {\hspace{.2cm} \includegraphics[width=.9\linewidth, trim={0.2cm 0.5cm 0.2cm 0.5cm}, clip]{imgs/lp_legend.pdf}}
    \subfloat[Maze Navigation]{\includegraphics[width=0.25\linewidth]{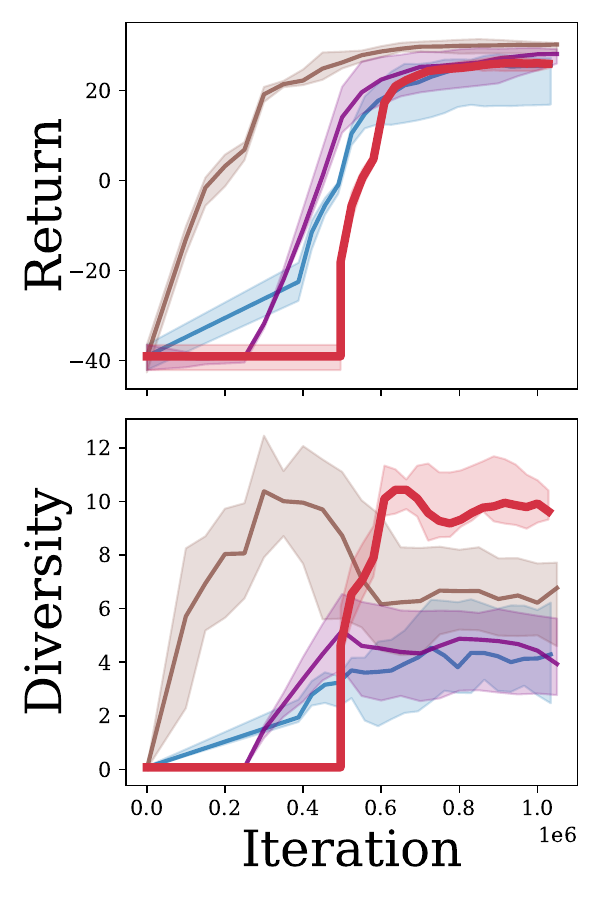}}
    \subfloat[Button Press]{\includegraphics[width=0.25\linewidth]{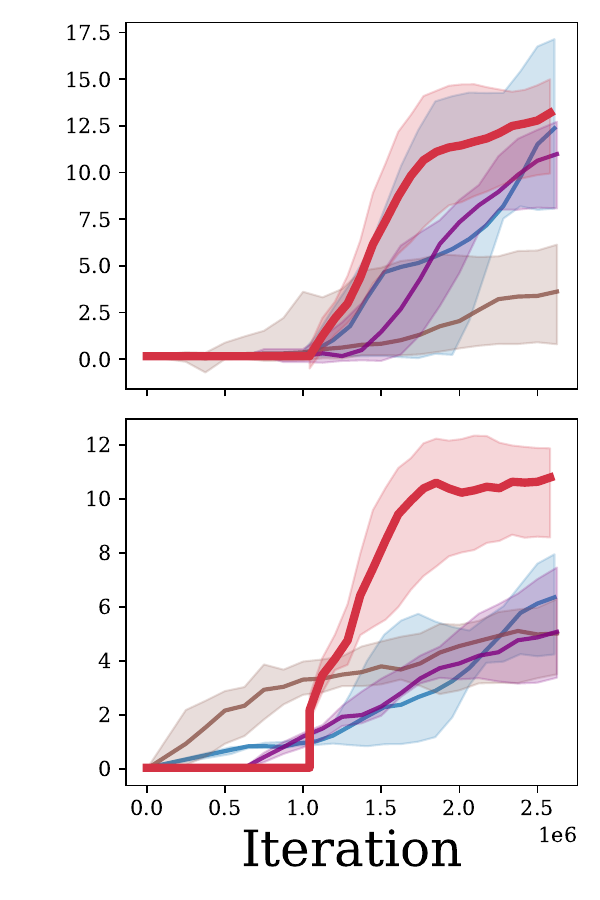}}
    \subfloat[Cube Push]{\includegraphics[width=0.25\linewidth]{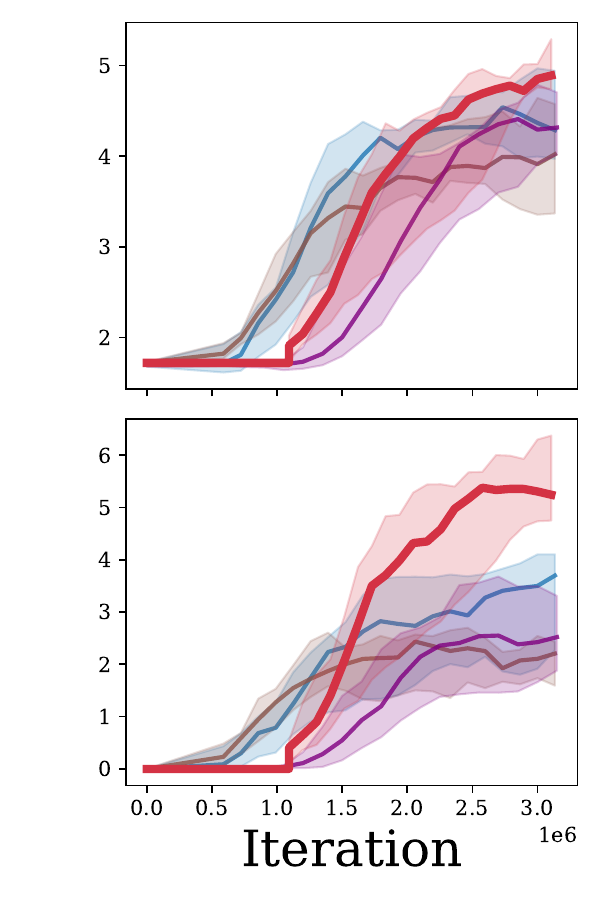}}
    \subfloat[Cube Flip]{\includegraphics[width=0.25\linewidth]{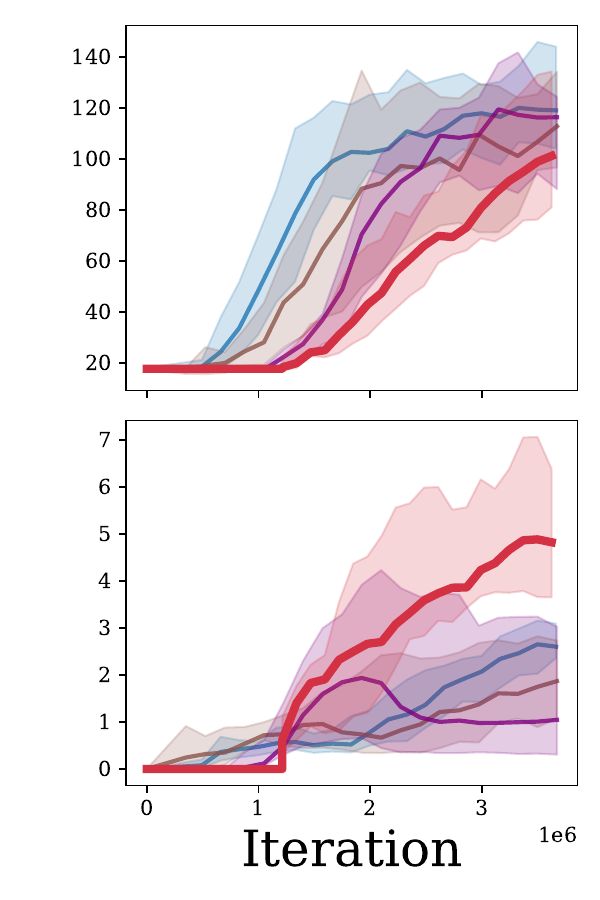}}
    \caption{Full learning curves. We report IQM and bootstrapped 95\%-CIs across 5 seeds.}
    \label{fig:learn_curves}
\end{figure}

We list the full learning curves in Fig.\ \ref{fig:learn_curves}.
These curves illustrate the evolution of both task performance and behavioral diversity over the course of training.
Overall, we observe strong final performance of our method.
A characteristic feature of our method is the CNS-based exploration during early environment interactions.
This is reflected by flat curves at the start, since policy learning is delayed until the second curriculum stage starts.
Yet, we can see that our method catches up quickly and surpasses all baselines in diversity.
The accompanying qualitative results are depicted in \Cref{fig:qual_res}.

\begin{wrapfigure}{R}{0.5\textwidth}
  \centering
  \includegraphics[width=\linewidth]{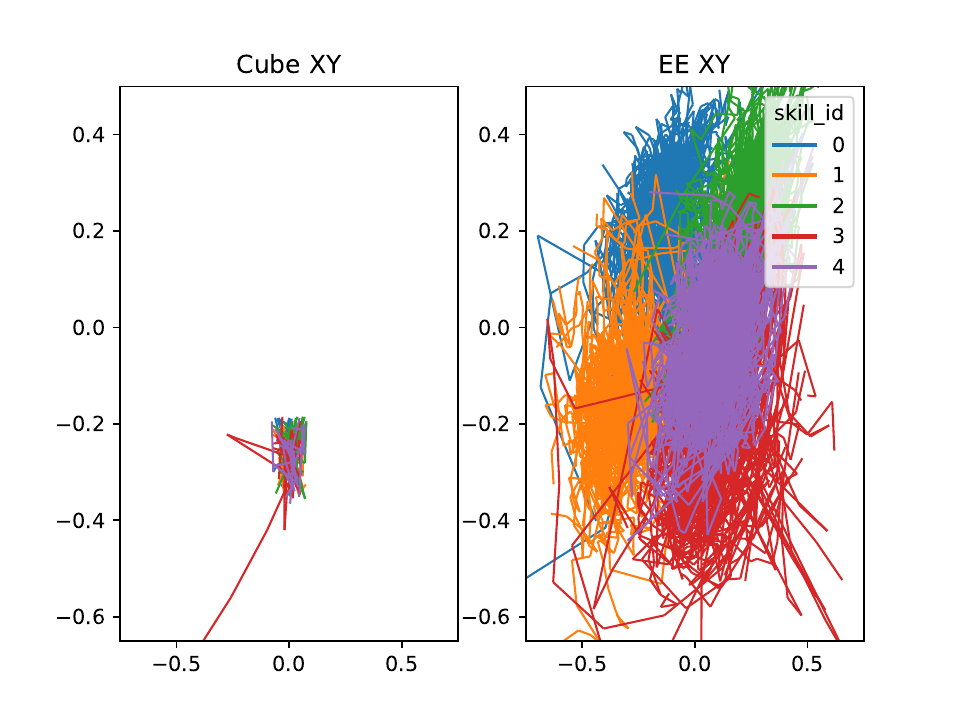}
  \caption{Exploration data of DIAYN in the cube push task. While endeffector poses are distinguishable between skills, object poses are not. The variation inside the box arises solely from non-deterministic resets. Only a single episode of skill 3 changed the object position.}
  \label{fig:diayn_data}
\end{wrapfigure}

\subsection{Diversity Objectives for Robot Manipulation}\label{secDivMan}
Our empirical results illustrate that diversity optimization in robot manipulation is a challenging problem.
In the following, we discuss the challenges that make this problem unique.

\paragraph{Contact Gradients are Challenging.}
In manipulation, the object state remains static unless a contact event occurs. 
Because these contacts are intermittent, i.e., continuously made and broken, the resulting gradients for both trajectory and policy optimization are notoriously sparse and discontinuous~\citep{suh2022differentiable, li2025drop}.
Since behavioral diversity in this context necessitates discovering different contact modes, the diversity objective itself is hard to optimize until the robot successfully manipulates the object~\citep{choi2024unsupervised}. 
This creates a restrictive dependency: meaningful diversity can only emerge once the robot has achieved a baseline level of task competence.
For this reason, the constrained diversity objective in \Cref{eq:domino} is so suitable to diversity maximization in manipulation.

\paragraph{Local Optima and Behavior Collapse.}
While the constrained diversity objective is designed for these scenarios, it faces a significant performance-diversity trade-off.
If the optimizer discovers a single successful strategy, e.g., a specific push, focusing on task performance can make the optimization susceptible to local optima.
In the following, attempting to find a new angle of attack to increase diversity often leads to a temporary drop in extrinsic rewards, causing the framework to revert to the known, high-reward mode.
Without an inductive bias like our CNS approach, these frameworks risk collapsing diversity in favor of stable task completion.

\paragraph{The Challenges of Discriminator-Based Diversity.}
The characteristics of the manipulation problem have implications for alternative exploration approaches too.
In particular, skill-based methods like DIAYN seem like ideal methods to discover diverse skills at first glance.
However, choosing the right features that the discriminator sees is very challenging for manipulation.
As shown in \Cref{fig:diayn_data}, when trained on full state observations, these methods often discover distinct end-effector trajectories that fail to interact with the object. 
This occurs because moving the arm through free space is a simpler and more effective way to satisfy the diversity discriminator than the high-risk, discontinuous approach of making contact. 
Conversely, if the discriminator is restricted to object features alone, it tends to overfit to spurious noise or non-deterministic resets during the early stages of training before any object-centric diversity has emerged~\citep{cathomen2025divide}.

\subsection{Full \texorpdfstring{$v_{\max}$}{v-max} Analysis}
\begin{figure}[bh]
    \centering
    \subfloat[Trajectory First]{
    \includegraphics[width=0.5\linewidth]{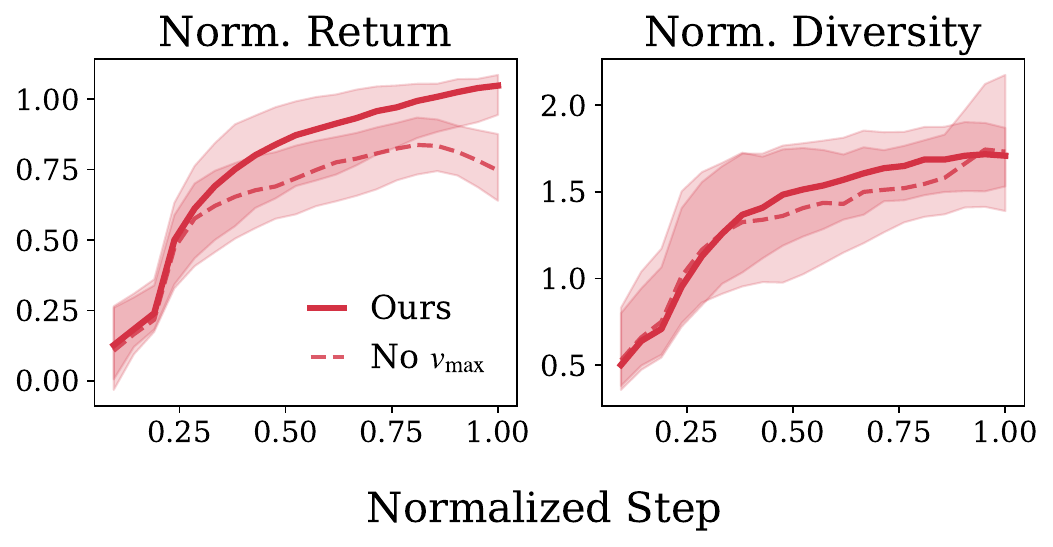}
    }
    \subfloat[Domino]{
    \includegraphics[width=0.5\linewidth]{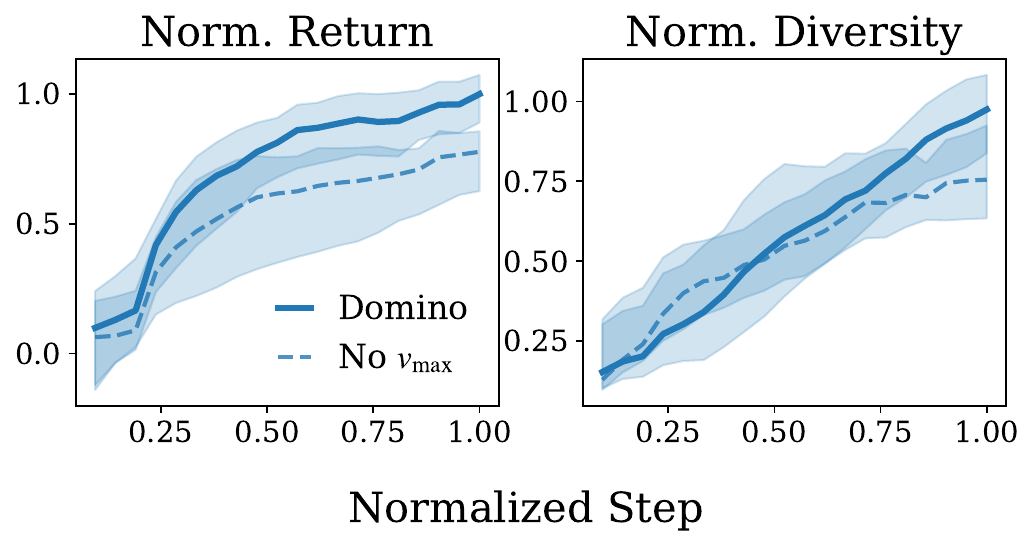}
    }
    \caption{The effect of the $v_{\max}$-trick on our curriculum method compared to its effect on Domino. The plots depict Domino-relative normalized performance over all tasks and 5 seeds. We report IQM and 95\% CIs. Using $v_{\max}$ to estimate the optimal MDP value improves policy returns while having no impact on diversity for both Trajectory First and standard Domino.}
    \label{fig:vmax}
\end{figure}

In \Cref{fig:vmax} we report additional results illustrating the effect of the $v_{\max}$-trick. 
The plots show that accurate estimation of the optimal value $v^*$ is crucial for high-performing policies.
In addition, we see that for Domino, the ablation of the $v_{\max}$-trick also yields lower diversity of the policy set.
When qualitatively inspecting the policies, we find that this is since they tend not to manipulate the object sufficiently well to induce object-level diversity in these cases.
These underline the significance of the $v_{\max}$-trick beyond our curriculum.
Hence we stress again that all results in the main paper use the $v_{\max}$ trick unless explicitly stated in the ablation study.

\subsection{CNS Analysis}\label{sec:cns_res}
\begin{figure}[t]
    \centering
    \includegraphics[width=.5\linewidth]{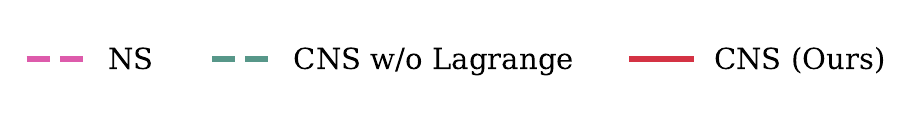}\\[-.5cm]
    \subfloat[Maze Navigation]{\includegraphics[width=0.25\linewidth]{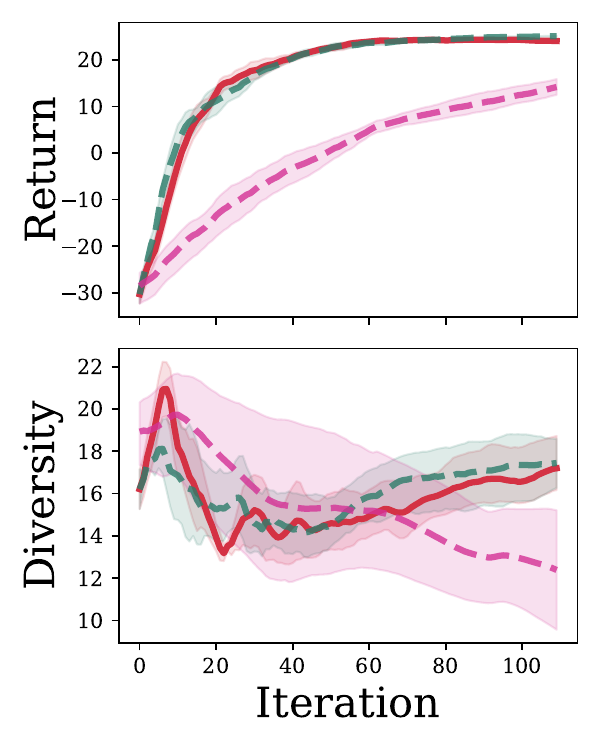}}
    \subfloat[Button Press]{\includegraphics[width=0.25\linewidth]{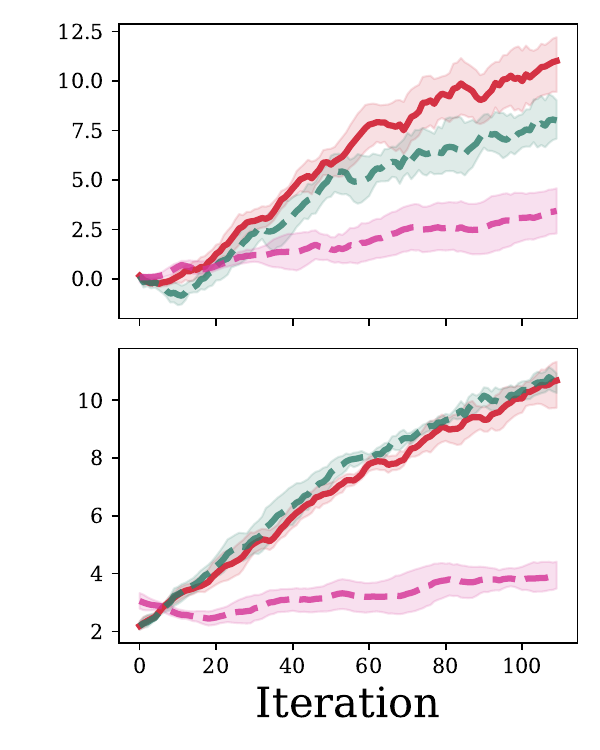}}
    \subfloat[Cube Push]{\includegraphics[width=0.25\linewidth]{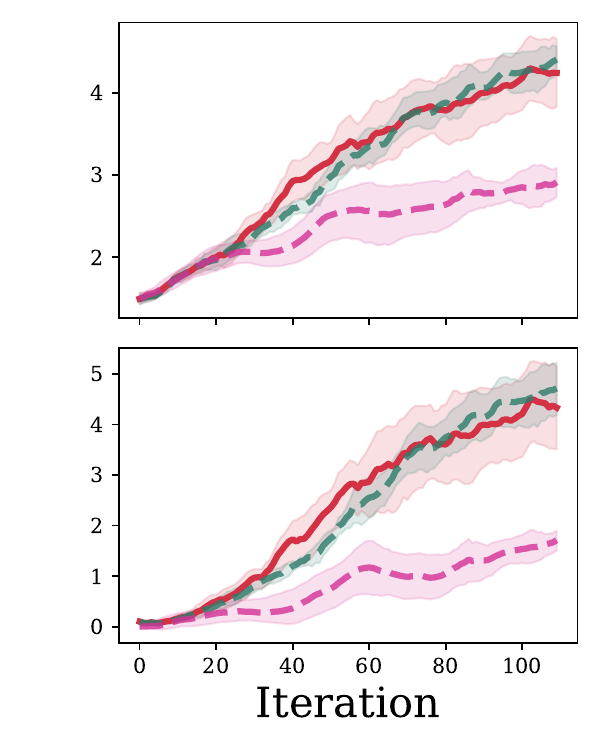}}
    \subfloat[Cube Flip]{\includegraphics[width=0.25\linewidth]{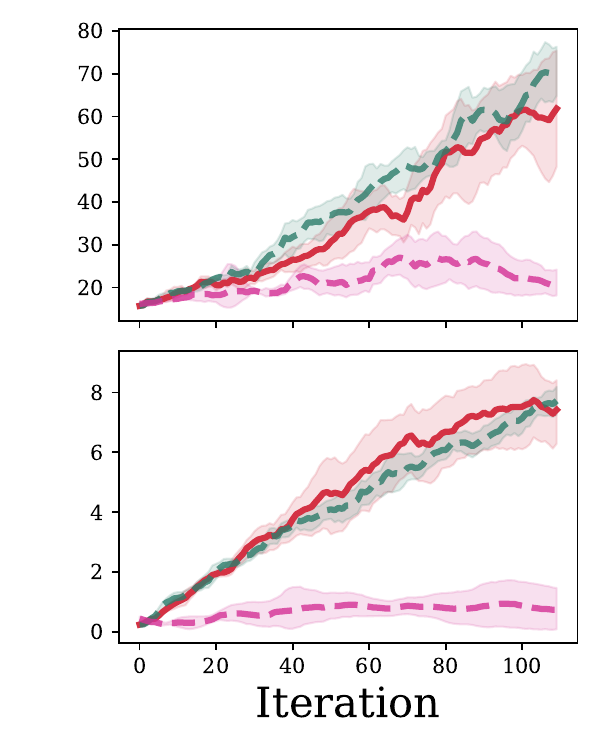}}
    \caption{CNS Analysis. Compared to standard NS CNS achieves higher returns and diversity. Further, we observe that Lagrange multipliers boost performance on the button press task, while being a more interpretable hyperparameter than a scalar mixing weight.}
    \label{fig:es_res}
\end{figure}

We analyze the performance of the proposed \textit{Constrained Novelty Search (CNS)} in the following.
We compare it against plain Novelty Search~\citep[NS]{conti2018improving} and an ablated version of CNS that does not use the dynamic Lagrange multipliers and instead relies on a hand-tuned scalar weight to trade extrinsic and intrinsic reward in \Cref{eq:cns}. 
The results in Fig.\ \ref{fig:es_res} show that CNS outperforms NS on all tasks in return and diversity.
This underlines that using non-isotropic search distributions for diversity optimization is a key component.
The original work that introduced NS for diversity maximization optimizes policies in neural network parameter space, which prohibits the usage of second-order-approximating methods like CMA-ES.
Our insight to instead optimize spline parameters permits us to improve the outcome of the behavior optimization considerably.

Further, we see that the introduction of the Lagrange multipliers does not reduce the performance of CNS.
This shows that CNS is a lightweight alternative to NS.
In contrast, we even observe higher returns for CNS on the button press task, which indicates the efficacy of the multipliers in enforcing the near-optimality constraint.
Instead of requiring a parameter search over the scaling of diversity and task reward, our method offers a simple and intuitive way to trade diversity against optimality.
In many tasks, being $75\%$ optimal has a clear intuition and depends solely on user preferences instead of meticulous hyperparameter tuning.
We therefore believe that CNS can make an impact beyond our curriculum as a diversity-seeking trajectory optimization method.

\section{Experimental Details}\label{sec:exp_dets}
\begin{figure}[t]
    \centering
    \subfloat[Maze Navigation]{
    \includegraphics[width=0.25\linewidth, page=1]{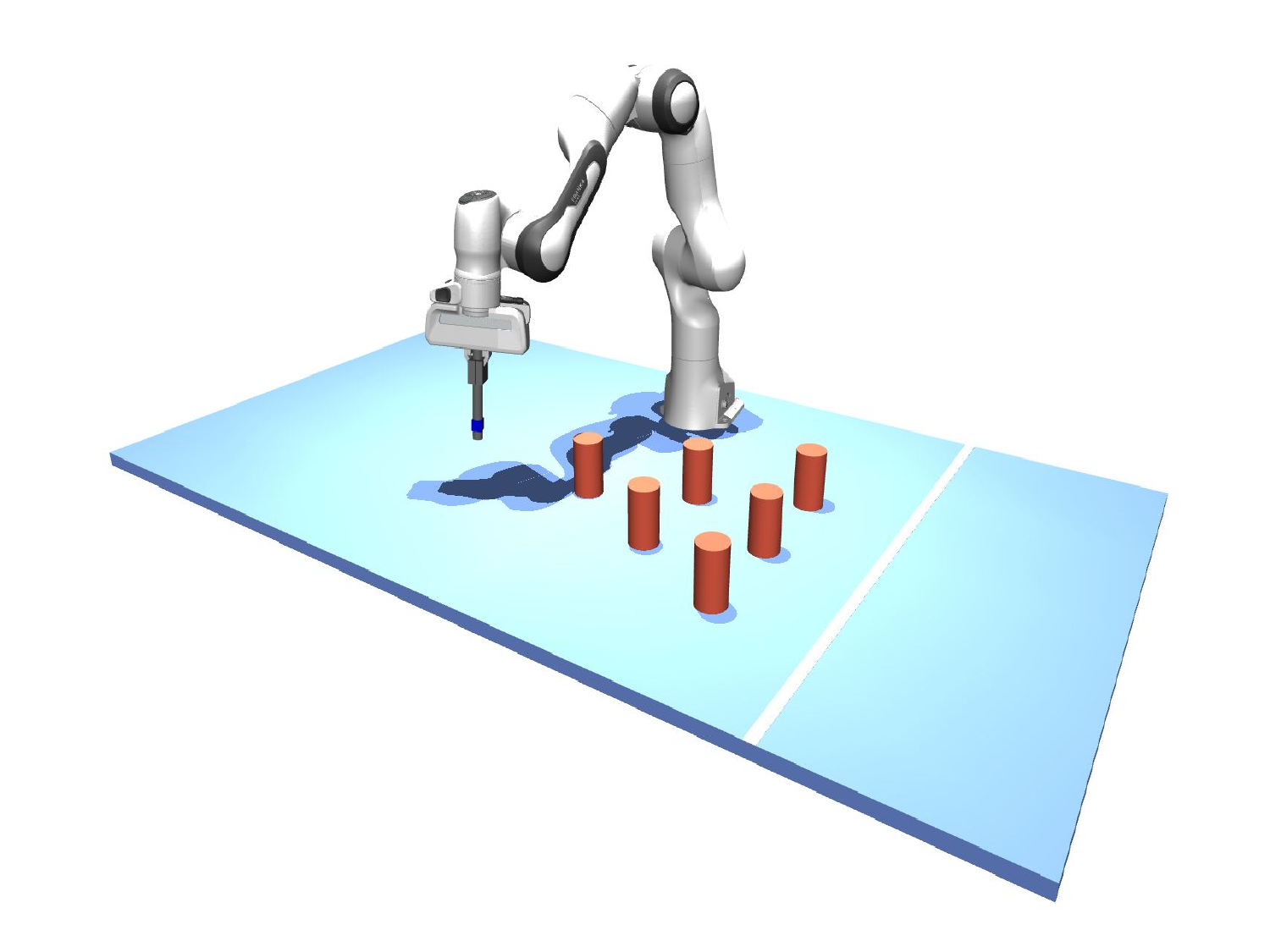}
    }
    \subfloat[Button Press]{
    \includegraphics[width=0.25\linewidth, page=4]{imgs/rlc_plot.pdf}
    }
    \subfloat[Cube Push]{
    \includegraphics[width=0.25\linewidth, page=3]{imgs/rlc_plot.pdf}
    }
    \subfloat[Cube Flip]{
    \includegraphics[width=0.25\linewidth, page=2]{imgs/rlc_plot.pdf}
    }
    \caption{The environments for our experiments. Each environment has a high degree of behavioral multimodality. For instance, objects can be manipulated in different ways and with different parts of the robot. Further, the navigation environment offers multiple paths that are near-optimal.}
    \label{fig:envs}
\end{figure}

Each experiment is repeated across 5 different seeds. 
Where applicable, we report the interquartile mean (IQM) across all 5 runs and bootstrapped 95\% confidence intervals in our plots.
We compute all statistics using numpy and scipy's \texttt{stats.bootstrap}.
In the following, we provide details about the environments and implementation that we used in this work.
All the experiments are performed on an internal cluster with two NVIDIA RTX PRO 6000 and eight NVIDIA A40 GPUs.
All environments are implemented in Mujoco \citep{todorov2012mujoco}. The tasks are depicted in \Cref{fig:envs}. Further details are listed below.

\subsection{Environments}

\paragraph{Maze Navigation.}
In this task, a robot must successfully navigate a rod that is attached to its gripper around 6 cylinder obstacles without collisions.
The agent is rewarded for minimizing the distance to the goal line and penalized for touching obstacles.
To avoid the trivial solution of moving over the top of all obstacles, we fix the $z$-position of the rod and use $xy$-endeffector position control.
Thus, the action space is $a \in [-1, 1]^2$, while the observation space consists of position and velocity information in the $xy$-plane, i.e., $s = [q, \dot{q}] \in \RRR^4$.
We train 10 different skills for this task.
The reward function that we use is the following:
\begin{equation*}
    r(s, a) = \beta_{\text{target}} (x_s - x_{\text{target}}) - \beta_{\text{coll}} \cdot \mathds{1}_{\text{coll}}(s) + x_{max}, \quad \beta_{\text{coll}} = 100;~\beta_{\text{target}} = \begin{cases}
        10, & \text{if } \mathrm{final}_t,\\
        1, & \text{otherwise.}
        \end{cases}
\end{equation*}
where $x_s$ denotes the $x$-coordinate of the rod, $x_{\text{target}}$ is the coordinate of the goal line, while $\mathds{1}_{\text{coll}}(s)$ is a collision checking function.
Further $x_{max}$ is the maximum $x$ coordinate that is admissible, which we use as offset to guarantee position rewards.
We additionally clip the step rewards to be in $[-1, 2]$ to improve training stability.
To further simplify the task, we bound the $xy$-positions to $[-4.5, 4.5]^2$, which we implement by clipping.
When training and evaluating the RL policies, we randomize the reset position by adding uniform jitter: $q_0 = q_{init} + \mathcal{U}[-0.5, 0.5]^2$.

\paragraph{Button Press.}
In this environment, an 8-DOF Panda robot is tasked to press a button in a box in front of it.
The agent is rewarded for approaching the button with any link and for pressing the button down.
To ensure safety, there are collision penalties for the robot links when touching any part of the box that is not the button:
\begin{align*}
    &r^{\mathrm{touch}}_t = 1 - \tanh \bigl(d_{\min,t}\bigr),\\
    &r^{\mathrm{press}}_t = z^{\mathrm{btn}}_0 - z^{\mathrm{btn}}_t,\\
    \intertext{Where $d_{\min,t}$ denotes the minimum distance between button and robot across all links and $z^{\mathrm{btn}}_t$ is the $z$-coordinate of the button at time $t$. Let $F^{\mathrm{coll}}_{i,t}\in\mathbb{R}^3$ be the measured force vector for collision with geometry $i$ at time $t$, and define the maximum force magnitude}
&F^{\max}_{\mathrm{coll},t} = \max_i \left\lVert F^{\mathrm{coll}}_{i,t}\right\rVert_2.\\
\intertext{The collision penalty is then}
&r^{\mathrm{coll}}_t = \tanh\!\left(\frac{F^{\max}_{\mathrm{coll},t}}{20}\right).
\intertext{Combining these terms, the total reward is defined as:}
&r_t = 0.001 r^{\mathrm{touch}}_t + \left( 2.5\, r^{\mathrm{press}}_t - 0.2\, r^{\mathrm{coll}}_t \right).
\end{align*}
The action space are joint poses $a \in [-1, 1]^8$, while the observation space are position and velocity information of robot and button as well as the contact forces between robot and button.
Specifically, the observation is defined as $o = [q, \dot{q}, p^{\text{btn}_t}, d_t, \bar{f}] \in \RRR^{66}$, where $q, \dot{q}$ are joint angles and velocities, $p_t^{\text{btn}} \in \RRR^3$ is the Cartesian button tip position, $d_t \in \RRR^7$ contains the distances between robot and its main frames and $\bar{f} \in 40$ contains the average contact force between all pairs of robot and environment bodies including the button.

We train 5 skills for this environment and measure diversity in the contact force and joint angle space.
In other words, we measure the contact forces between each link of the robot and the button as well as the difference in joint angles to estimate behavioral diversity.
When training and evaluating the RL policies, we randomize the reset pose of the robot by applying uniform angle transformations in up to : $q_0 = q_{init} + \mathcal{U}[-0.17, 0.17]^8$.

\paragraph{Cube Push.}
In this task, an 8-DOF Panda robot is tasked to push a cube as far as possible.
The agent is only rewarded for maximizing the distance between the cube's center of mass and its initial position.
Concretely, the reward combines a small positive term for touching the cube with the endeffector, and small acceleration and collision penalties to prevent the robot from hitting the table.
The full reward is defined as:
\begin{align*}
    &r_t = 5.0r^{\mathrm{dist}}_t + 0.01 r^{\mathrm{touch}}_t - 0.001\, r^{\mathrm{acc}}_t - 0.005\, r^{\mathrm{coll}}_t.\\
    \intertext{The components are defined as follows:}
    &r^{\text{coll}}_t = \mathds{1}_{\text{coll}} + \tanh\!\left(\frac{F^{\max}_{\mathrm{coll},t}}{20}\right),\\
    &r^{\mathrm{touch}}_t = 1 - \tanh\!\left(\left\lVert p^{\mathrm{cube}}_t - p^{\mathrm{ee}}_t \right\rVert^2_2\right),\\
    &r^{\mathrm{acc}}_t = \operatorname{clip} \left[\left\lVert \ddot q_t\right\rVert_2^2;~0, 1\right],\\
    &r^{\mathrm{dist}}_t = \delta_t - \delta_{t-1},\text{ where  }\delta_t=\lVert p^{\mathrm{cube}}_{t,xy} - p^{\mathrm{cube}}_{0,xy}\rVert_2.
\end{align*}
This task is challenging because of the deceptive rewards in the contact-rich manipulation setting.

The action space are endeffector poses $a \in [-1, 1]^10$, which we represent as 3D position and 6D rotation following \citet{zhou2019continuity}.
The observation consists of the endeffector and cube poses, the relative transformation between the two, the cube velocities, as well as the previous action.
Concretely, the observation is $o = [P^{\text{cube}}, P^{\text{EE}}, \dot{P}^{\text{cube}}, \dot{P}^{\text{EE}}, \left(R^{\mathrm{ee}}_t\right)^\top \!\left(p^{\mathrm{cube}}_t - p^{\mathrm{ee}}_t\right), q_t, \dot q_t] \in \RRR^{43}$
We train 5 skills for this environment and measure diversity using the cube position in the $xy$-plane.
When training and evaluating the RL policies, we randomize the reset pose of the robot by applying uniform angle transformations in up to : $q_0 = q_{init} + \mathcal{U}[-0.17, 0.17]^8$. In addition, we randomize the initial cube pose by sampling position and rotation offsets uniformly. 
Positions are sampled from $\mathcal{U}[-5, 5]^2$, denoting $xy$-deltas in centimeters. 
Rotations are sampled from $\mathcal{U}[-20, 20]$, denoting rotation around $z$ in degrees.

\paragraph{Cube Flip.}
In this task, an 8-DOF Panda robot is tasked to flip a cube.
The agent is rewarded for maximizing the $z$-component of the red face normal of the cube.
Concretely, the reward combines a small positive term for touching the cube with the endeffector, and the angle reward.
The full reward is defined as:
\begin{align*}
    &r_t = r^{\mathrm{flip}}_t + 0.1\, r^{\mathrm{touch}}_t,\\
    \intertext{where}
    &r^{\mathrm{flip}}_t = \operatorname{clip}\!\left[n^{\mathrm{red}}_{t}\cdot \hat z;~0,1\right], \\
    &r^{\mathrm{touch}}_t = 1 - \tanh\!\left(\left\lVert p^{\mathrm{cube}}_t - p^{\mathrm{ee}}_t \right\rVert_2\right).
\end{align*}
Here, $n^{\mathrm{red}}_{t}$ denotes the world-frame normal of the cube's red face, and $\hat z$ is the world $z$-axis.

The action space follows the same control interface as in \paragraph{Cube Push}, i.e., we use end-effector pose control (3D position and 6D rotation, plus a gripper command).
The observation consists of the end-effector and cube poses, the relative cube position expressed in the end-effector frame, the robot joint angles and velocities, the cube linear and angular velocities, as well as contact- and dynamics-related signals.
Concretely, the observation is
\begin{align*}
o_t = [&\,p^{\text{EE}}_t, R^{\text{EE}}_{6D,t}, p^{\text{cube}}_t, R^{\text{cube}}_{6D,t},
\left(R^{\mathrm{EE}}_t\right)^\top \!\left(p^{\mathrm{cube}}_t - p^{\mathrm{EE}}_t\right),\\
&\, q_t, \dot q_t, v^{\text{cube}}_t, \omega^{\text{cube}}_t, f^{\text{coll}}_t, w^{\text{ext}}_t ] \in \RRR^{55},
\end{align*}
where $R_{6D}$ denotes the 6D rotation representation, $f^{\text{coll}}_t\in\RRR^6$ are aggregated contact force magnitudes, and $w^{\text{ext}}_t\in\RRR^6$ is the external wrench acting on the cube.
When training and evaluating the RL policies, we randomize the reset pose of the robot by sampling joint perturbations proportional to each joint's range, and randomize the initial cube pose by sampling position offsets just as in the cube push environment.

\subsection{Implementation Details.}
We implement all algorithms in JAX \citep{bradbury2018jax}.
The code for all RL agents is based on Domino \citep{zahavy2023discovering}, a public implementation thereof \citet{grillotti2024quality}, and the STOIX ecosystem \citep{toledo2024stoix}.
We follow \citet{zahavy2023discovering} in the choices of all hyperparameters for Domino with exceptions detailed below.
For all baselines, we use ground truth observations as state features, since they are low-dimensional and should thus yield the best performances \citep{zahavy2023discovering, g2024discovering}.
For a full list of hyperparameters, we refer to \Cref{table:hyperparams}.

\paragraph{Initialization.}
Since we initialize Domino from prior data, we adapt the initialization of the feature EMAs.
The running estimates of the state features are not initialized with $\phi^{avg}= \bar{1}/f$ for features in $\RRR^f$.
Instead, we use the mean of the maximum likelihood solutions from the CNS.
In other words, for each CMA-ES population that we run, we select the resulting trajectory parameters, roll out an additional trajectory from them and use the expected features over this trajectory as initial estimate of the features per skill.
We find that this initialization is provides better results than the uniform initialization from Domino.
For the values however, we follow Domino in using a zero initialization for all skills instead of using the expected reward from the final parameter rollouts.
This is because such an initialization would overestimate the capacities of the current policies and thus only optimize diversity from the very beginning of training following Eq.\ \ref{eq:domino}.

\paragraph{Network Architectures.}
As stated in \Cref{sec:learning} we follow design choices from prior work in using layer normalization and observation normalization.
We use the Simba architecture \citep{lee2024simba} for both actor and critic.
For further regularization, we perform critic ensembling and use separate heads for extrinsic and intrinsic values.
All networks are optimized using Adam \citep{kingma2015adam}.
To obtain similarly scaled extrinsic and intrinsic targets, we standardize rewards using running statistics, as recommended in \citet{lee2024simba}.

\paragraph{Baseline Implementation Details.}
We implement DIAYN following the original implementation in \cite{eysenbach2018diversity}, but port the code to JAX.
We use a uniform prior over skills and use a 2-layer MLP with 256 units each for the discriminator.
For parameter noise exploration, we implement NoisyNet~\citep{fortunato2018noisy} for both actor and critic networks.
Specifically, we use the public JAX implementation from \citet{toledo2024stoix} and disable parameter noise for inference and evaluation.
For all baselines, we use the complete set of stabilization mechanisms that we discuss in the main section to guarantee a fair comparison.

\paragraph{Constrained Novelty Search.}
We implement CNS based on the CMA-ES implementation in evosax \citep{evosax2022github}.
Before combining extrinsic and intrinsic fitness, we normalize both values within each subpopulation.
Further, we use simple gradient descent to update the Lagrange multipliers.
For higher optimization stability, we only update these parameters every iteration, but then perform 200 steps of gradient descent.
To prevent gradient saturation due to the usage of sigmoids on the Lagrange multipliers, we bound them to make sure that they remain in a reasonable range.
Similar to Domino, we also fix the first Lagrange multiplier to $1$, so we can estimate $v^*$ based on this population.
We list the CNS hyperparameters in \Cref{table:hyperparams}.

\paragraph{Hyperparameters.}
To guarantee a fair comparison of methods, we tune key hyperparameters for our method and Domino separately.
Specifically, we use optuna~\citep{akiba2019optuna} for Bayesian optimization with a budget of 25 runs per method.
The parameters that we tune are the EMA weights $\alpha_\phi, \alpha_v$, the discount factor $\gamma$, the critic learning rate (we use the same for the agent), and the Lagrange multiplier learning rate.
For extensions of Domino, i.e., Domino+Diayn, and Domino+Parameter Noise we use the Domino hyperparameters.
The full list of hyperparameters can be found in \Cref{table:hyperparams}.

\newcommand{\rowlab}[1]{\rotatebox{90}{\bfseries\shortstack[c]{#1}}}

\begin{figure}
  \centering

  \begin{minipage}[t]{0.72\linewidth}
    \centering

    \newcommand{\labw}{0.07\linewidth}
    \newcommand{\imgw}{0.93\linewidth}

    \subfloat[Button Press]{%
      \resizebox{0.95\linewidth}{!}{%
        \begin{tabular}{@{}>{\centering\arraybackslash}m{\labw} m{\imgw}@{}}
          \rowlab{Domino}         & \includegraphics[width=\linewidth]{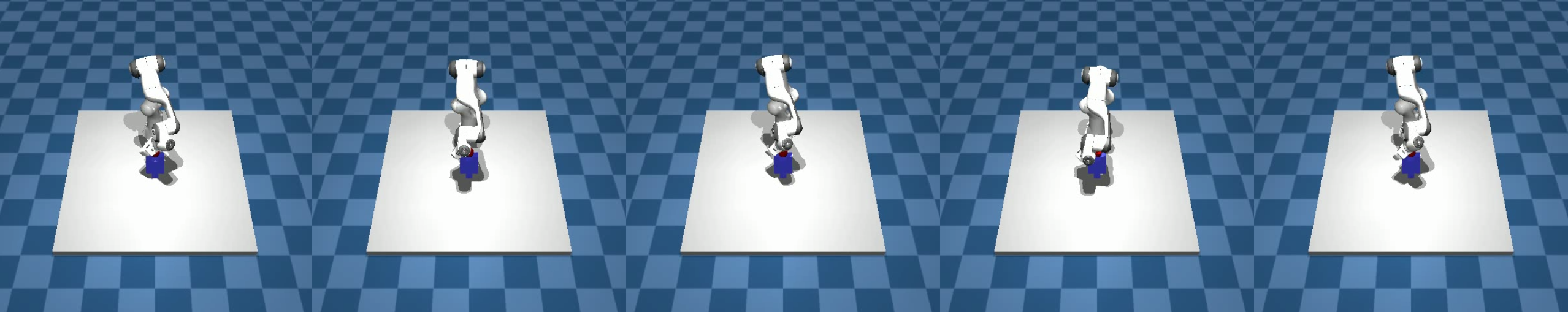}\\ 
          \rowlab{Domino\\+Pnoise}  & \includegraphics[width=\linewidth]{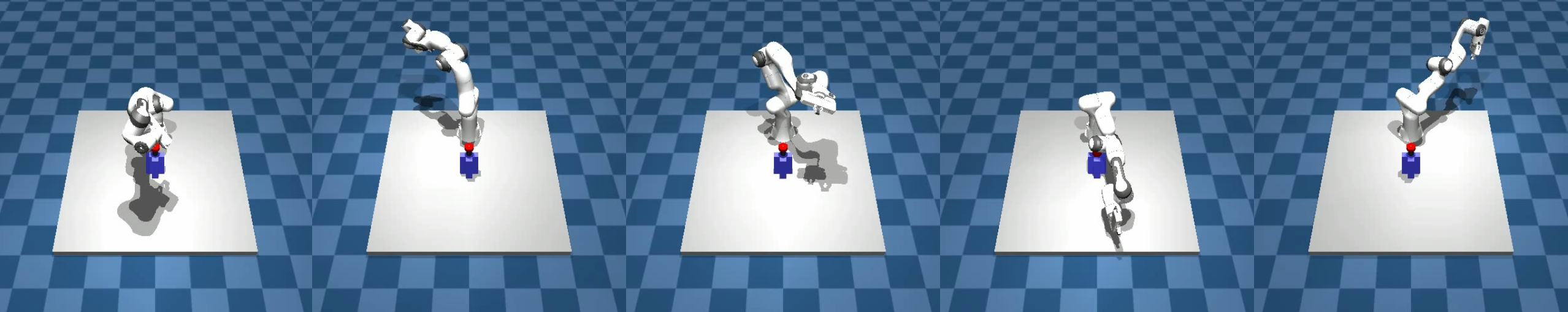}\\
          \rowlab{Domino\\+DIAYN}   & \includegraphics[width=\linewidth]{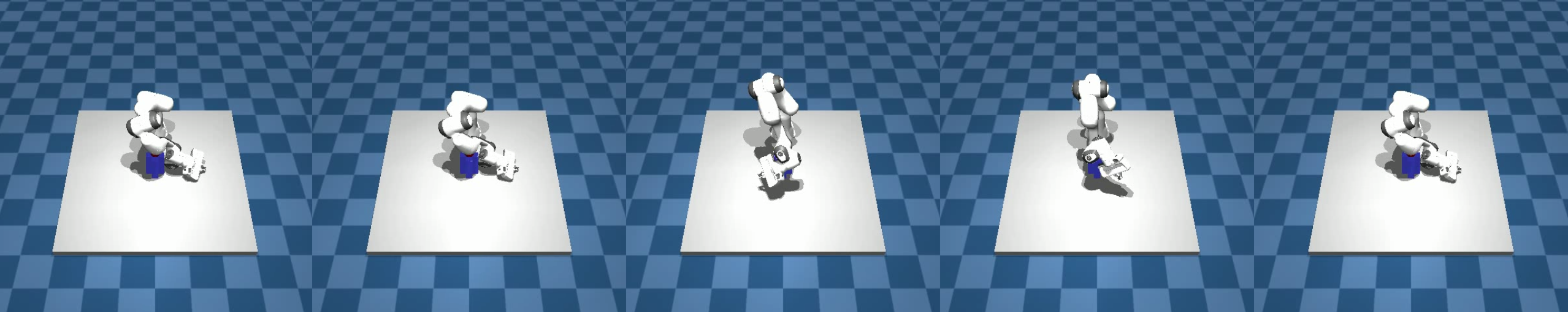}\\
          \rowlab{Ours}           & \includegraphics[width=\linewidth]{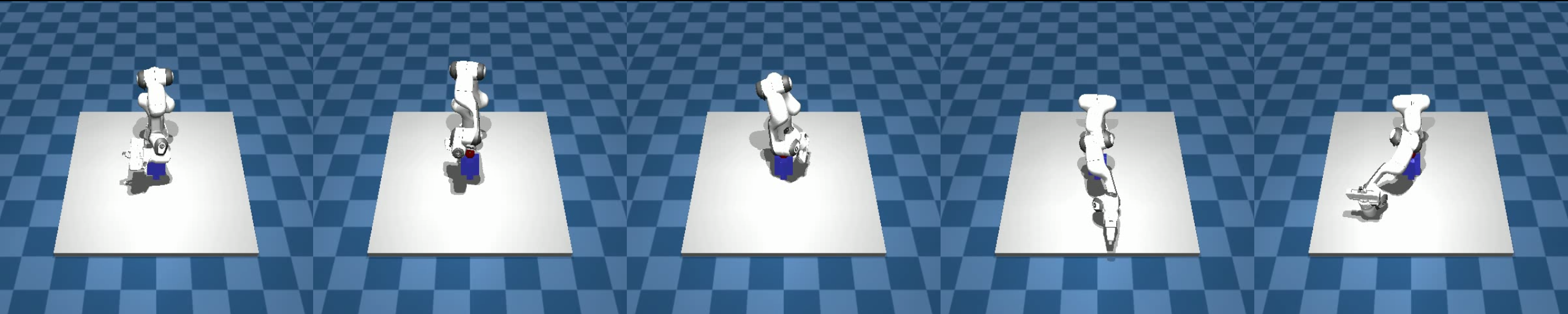}\\
        \end{tabular}%
      }%
    }\\[2pt]

    \subfloat[Cube Push]{%
      \resizebox{0.95\linewidth}{!}{%
        \begin{tabular}{@{}>{\centering\arraybackslash}m{\labw} m{\imgw}@{}}
          \rowlab{Domino}         & \includegraphics[width=\linewidth]{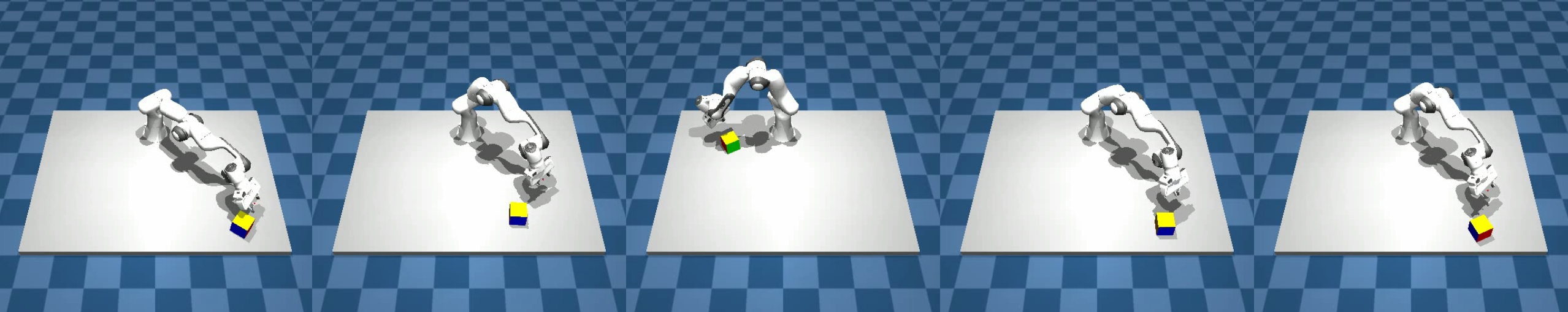}\\
          \rowlab{Domino\\+Pnoise}  & \includegraphics[width=\linewidth]{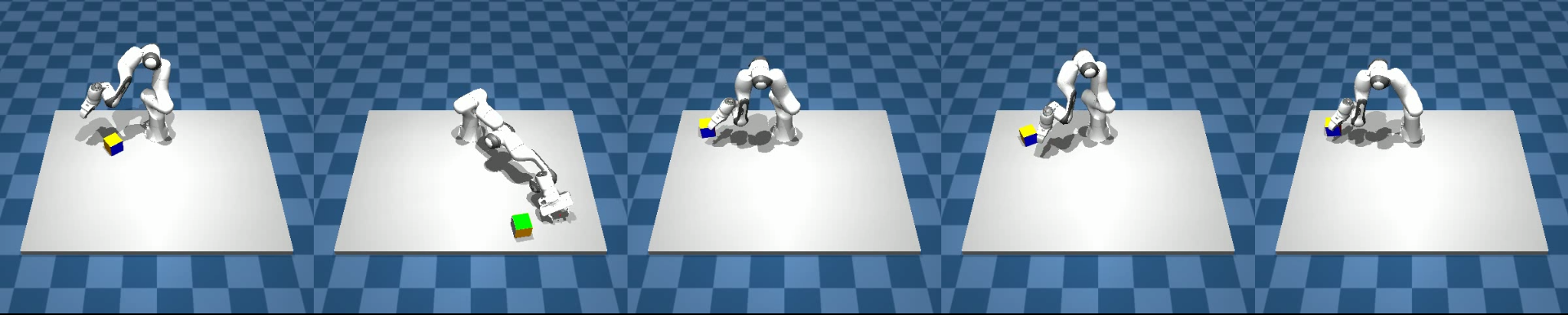}\\
          \rowlab{Domino\\+DIAYN}   & \includegraphics[width=\linewidth]{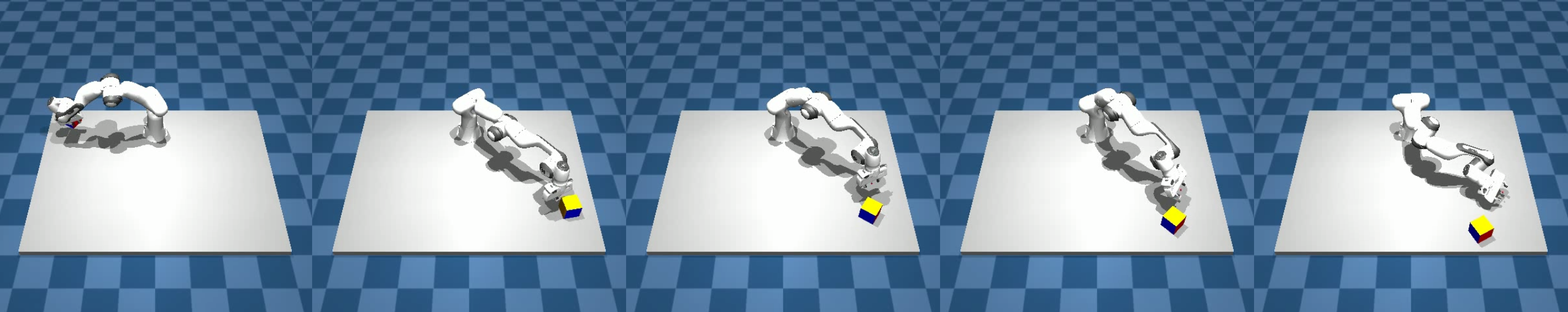}\\
          \rowlab{Ours}           & \includegraphics[width=\linewidth]{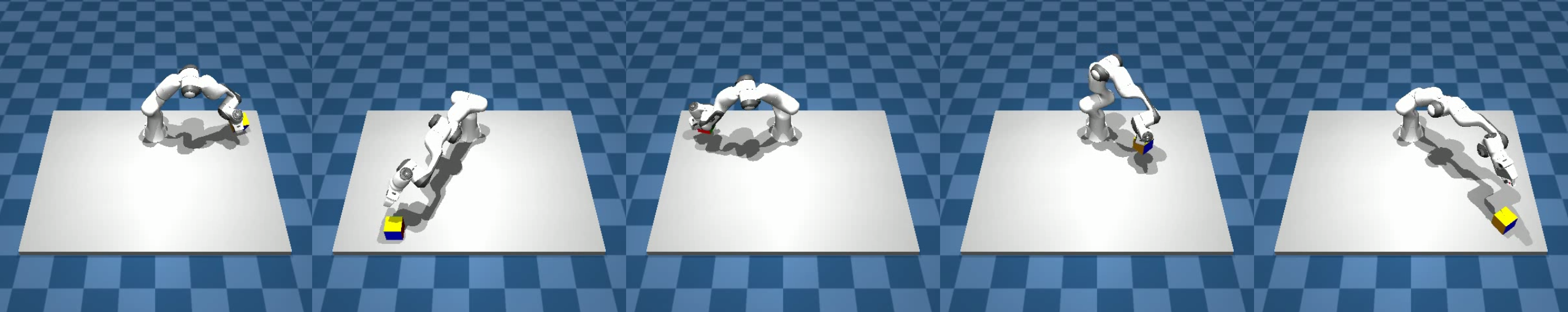}\\
        \end{tabular}%
      }%
    }\\[2pt]

    \subfloat[Cube Flip]{%
      \resizebox{0.95\linewidth}{!}{%
        \begin{tabular}{@{}>{\centering\arraybackslash}m{\labw} m{\imgw}@{}}
          \rowlab{Domino}         & \includegraphics[width=\linewidth]{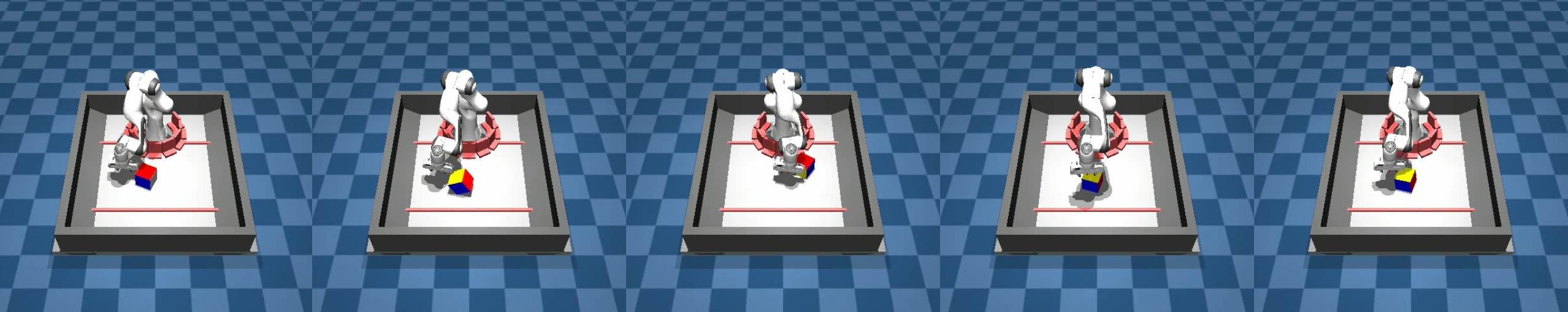}\\
          \rowlab{Domino\\+Pnoise}  & \includegraphics[width=\linewidth]{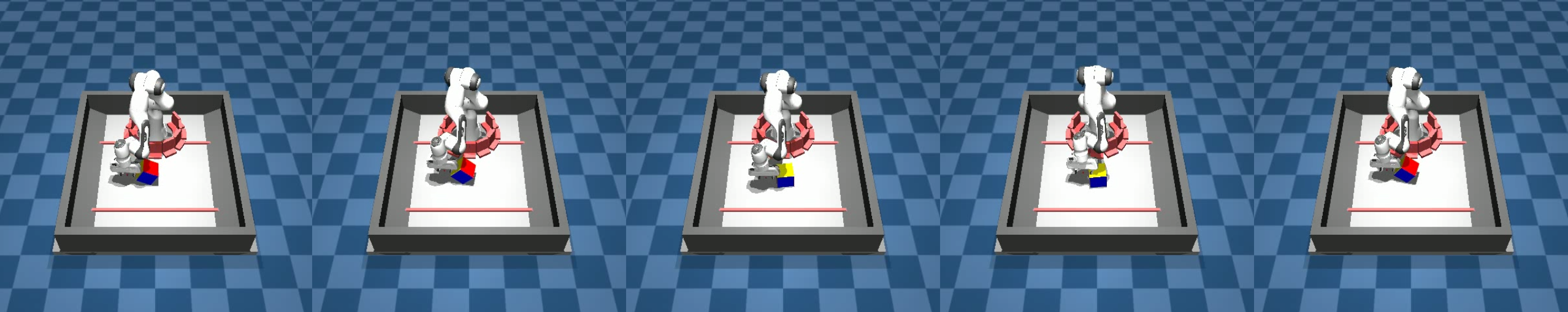}\\
          \rowlab{Domino\\+DIAYN}   & \includegraphics[width=\linewidth]{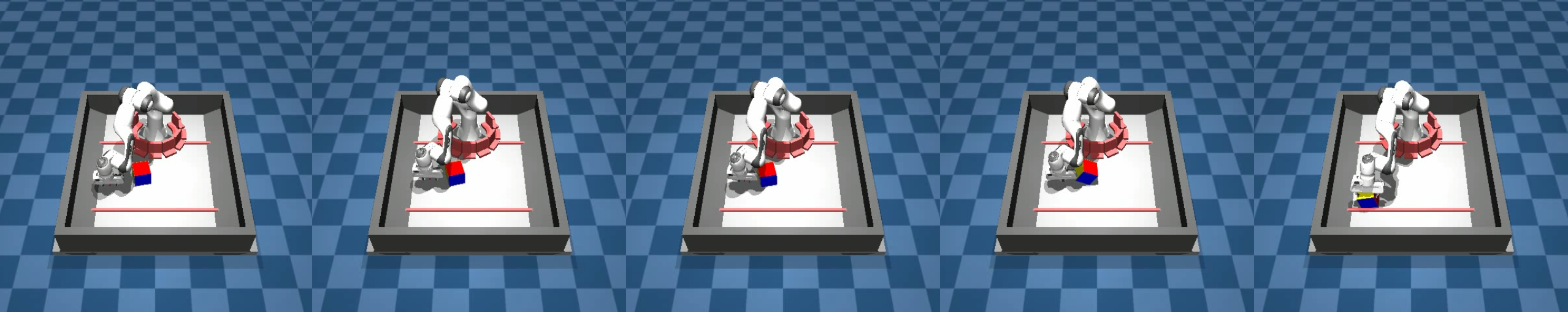}\\
          \rowlab{Ours}           & \includegraphics[width=\linewidth]{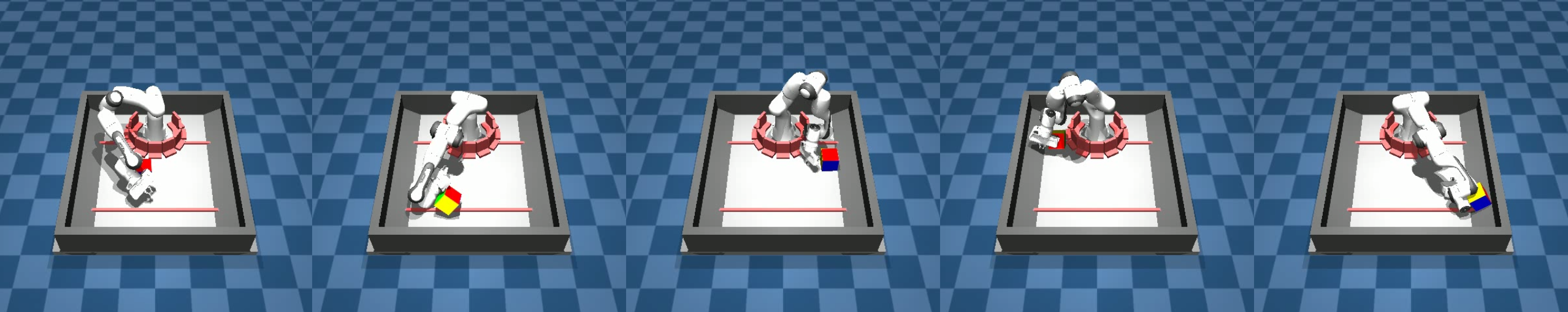}\\
        \end{tabular}%
      }%
    }
  \end{minipage}\hfill
  \begin{minipage}[t]{0.26\linewidth}
    \captionsetup{type=figure} 
    \captionof{figure}{Qualitative results. An extended version demonstrating the behavioral diversity our approach finds. For the cube push, we discover policies that push in different directions. For the button press, the discovered policies use different parts of the robot to press, performing whole-body manipulation. For the cube flip, we discover policies that use different parts of the environment as well as different force profiles to flip the cube.}
    \label{fig:qual_res}
  \end{minipage}
\end{figure}

\begin{table}
\centering
\begin{tabular}{@{} l  c  c  c  c @{}}
\toprule
\textbf{} & \textbf{Maze Navigation} & \textbf{Button Press} & \textbf{Cube Push} & \textbf{Cube Flip}\\
\midrule
\midrule
\multicolumn{3}{@{}l}{\itshape Environment Details} \\[3pt]
\quad Observation size      & 4  &  66 & 43 & 55\\
\quad Action size           & 2  &  10 & 8 & 10 \\
\quad Episode length        &  100  &  200 & 200 & 200 \\
\quad Num.\ env.\ steps         &  1\,000\,000  & 2\,500\,000 & 3\,000\,000 & 3\,500\,000  \\
\quad Num.\ skills         &  10  &  5 & 5 & 5 \\
\quad Optimality ratio $\alpha$   & 0.8 & 0.5 & 0.8 & 0.5 \\
\midrule
\midrule
\multicolumn{3}{@{}l}{\itshape RL Parameters (shared across methods)} \\[3pt]
\quad Num.\ envs            &  32  & 32  & 32 & 32 \\
\quad Update batch size            &  256  & 256  & 256 & 256 \\
\quad $\HH_{target}$         &  $\dim \AA / 2$  &  $\dim \AA / 2$ &  $\dim \AA/2$ & $\dim \AA/2$ \\
\quad Optimizer            &  Adam  &  Adam & Adam & Adam\\
\quad Polyak weight         &  5e-3 &  5e-3 & 5e-3 & 5e-3 \\
\quad Num.\ critics         &  10  &  10  & 10 & 10 \\
\quad Critic subset size         &  2  &  2 & 2 & 2  \\
\quad Critic hidden depth            &  4  &  4 & 4 & 2\\
\quad Critic hidden size           &  64  &  64 & 64 & 256\\
\quad Actor hidden depth            &  4  &  4 & 4 & 2\\
\quad Actor hidden size           &  64  &  64 & 64 & 256\\
\quad Learn.\ Rate Temperature            &  3e-4  &  3e-4 & 3e-4 & 3e-4 \\
\quad Buffer size               & 1e6 & 1e6 & 1e6 & 2e6 \\
\midrule

\multicolumn{3}{@{}l}{\itshape Trajectory First Parameters} \\[3pt]
\quad Learn.\ Rate Critic            &  3e-4  &  3e-4 & 3e-4 & 5e-4\\
\quad Learn.\ Rate Actor            &  3e-4  &  3e-4 & 3e-4 & 5e-4\\
\quad Learn.\ Rate Lagrange            &  5e-4  &  1e-4 & 3e-4 & 1e-3 \\
\quad Discount & 0.975  &  0.98 & 0.975 & 0.985 \\
\quad Policy update freq.\  &  4  &  4 & 2  & 4\\
\quad Critic update freq.\  &  4  &  4 & 8 & 4\\
\quad EMA weight $\alpha_{\phi}$         &  0.995  &  0.9994 & 0.995 & 0.9999 \\
\quad EMA weight $\alpha_{v}$         &  0.992  &  0.996 & 0.992 & 0.999 \\

\midrule
\multicolumn{3}{@{}l}{\itshape Baseline Parameters} \\[3pt]
\quad Learn.\ Rate Critic            &  1e-3  &  2e-4 & 6e-4 & 3e-4\\
\quad Learn.\ Rate Actor            &  1e-3  &  2e-4 & 6e-4 & 3e-4\\
\quad Learn.\ Rate Lagrange            &  5e-4  & 3e-4 & 6e-4 & 1e-3 \\
\quad Discount & 0.95  &  0.99 & 0.95 & 0.985 \\
\quad Policy update freq.\  &  4  &  2 & 2  & 4\\
\quad Critic update freq.\  &  4  &  2 & 8 & 4\\
\quad EMA weight $\alpha_{\phi}$         &  0.999  &  0.994 & 0.995 & 0.995 \\
\quad EMA weight $\alpha_{v}$         &  0.99  &  0.994 & 0.992 & 0.99 \\
\midrule
\multicolumn{3}{@{}l}{\itshape CNS Parameters} \\[3pt]
\quad Num.\ iterations        &  110  &  110 & 110 & 120 \\
\quad Subpopulation size         &  4  &  8  & 8 & 8 \\
\quad Elite ratio         &  0.5  &  0.25 & 0.25 & 0.25\\
\quad Init.\ $\sigma$         &  0.6  &  0.6 & 0.6 & 0.6 \\
\quad Num.\ spline controls  & 5 & 4 & 4 & 4\\

\bottomrule
\end{tabular}
\caption{RL Hyperparameter Overview}
\label{table:hyperparams}
\end{table}

\end{document}